\documentclass[runningheads]{article}
\usepackage{graphicx}
\usepackage{amsfonts}
\usepackage{amsmath}
\usepackage{authblk}
\begin{document}
\title{State of the Art in Artificial Intelligence applied to the Legal Domain}
%\title{NLP Applied To Portuguese Consumer Law\thanks{Supported by INCM (Imprensa Nacional Casa da Moeda).}}
%
%\titlerunning{Abbreviated paper title}
% If the paper title is too long for the running head, you can set
% an abbreviated paper title here
%

\author[3,2]{João Dias\thanks{joao.dias@gaips.inesc-id.pt}}
\author[1,2]{Pedro A. Santos}
\author[1,2]{Nuno Cordeiro}
\author[1,2]{Ana Antunes}
\author[1,2]{Bruno Martins}
\author[4,2]{Jorge Baptista}
\author[5]{Carlos Gonçalves}
\affil[1]{Instituto Superior Técnico, Universidade de Lisboa}
\affil[2]{INESC-ID}
\affil[3]{Faculdade de Ciências e Tecnologia, Universidade do Algarve}
\affil[4]{Faculdade de Ciências Sociais e Humanas, Universidade do Algarve}
\affil[5]{Imprensa Nacional Casa da Moeda}
%
%\authorrunning{N. Cordeiro et al.}
% First names are abbreviated in the running head.
% If there are more than two authors, 'et al.' is used.
%
%\institute{INESC-ID and Imprensa Nacional Casa da Moeda}
%
\maketitle              % typeset the header of the contribution
\begin{abstract}
    While Artificial Intelligence applied to the legal domain is a topic with origins in the last century, recent advances in Artificial Intelligence are posed to revolutionize it. This work presents an overview and contextualizes the main advances on the field of Natural Language Processing and how these advances have been used to further the state of the art in legal text analysis.
\end{abstract}

\section{Introduction}
Every country in the world bases its inner workings in a complex juridical system. For instance, in Portugal, the Official Portuguese Gazette (Diário da República) is the venue for the publication of all laws and norms of the Portuguese Republic. More than a million and a half juridic acts are currently active, and more than a thousand new ones are published in Diário da Républica Electrónico (DRE) every month. This online resource provides access to all of the Portuguese legislation, as well as services that allow citizens to find the laws and norms that they look for.

It is clearly important that this type of information is easily accessible to all citizens given that there are rights they have and obligations they must abide to. When accessing legislation search, some citizens may have difficulties in finding results to a search, given the level of language that is used. Regular citizens normally use Natural Language (NL) to introduce a search query. The problem is that laws and legal norms contain language that --- despite being technically NL --- uses specialized terminology and sentence construction. This creates a double barrier to even a simple search. On one hand the system must understand the user query and relate it to the correct piece of legislation. On the other hand, the returned text is not easily understood by regular users, due to its formality.

The logical character of legal texts, norms, procedures and argumentation suggests that formal logic can function as a model capable of representing legal rules and situational facts. Thus, it would be possible to help the citizen with inference based on the application of rules to facts. But, for such a system to work the ``natural language barrier" we mentioned above must be overcome several times. First, it is necessary to translate the legal norms from NL into some form of machine-compatible semantic or logical representation. This difficulty becomes a real barrier due to several peculiarities of NL and the legal language. Then also the user query must be translated from NL into a semantic or logic representation. And finally, after the logical inference has been performed by the system, the ``natural language barrier" has to be transposed a third time, to generate a user-friendly explanation or procedure description for the lay person.

Revolutionary advances in Natural Language Processing have very recently changed the landscape, creating scalable methods to analyse large corpora of text. They have started also to be applied in the legal domain, and open new paths to cross the barrier referred above. Many difficulties and open problems remain, but it seems that in the next few years we will be able to create new tools with which to develop innovative application to help citizens interface with the Law.

This paper presents an overview, contextualization and analysis of the state of the art in the area of Artificial Intelligence and Law, of Natural Language Processing (NLP) and its application to the legal field, with the goal of determining how to best leverage the recent advancements to improve the search services provided by DRE to users. 

%When looking for applications of the legislation area, we will focus on two main subareas, which have achieved interesting results in the last years, Legal Information Retrieval and Semantic Analysis for Automated Compliance Checking.  

This document is organized as follows. In the next section we offer a brief overview of the subfield of Artificial Intelligence applied to the legal domain. In the third section, Advances of Deep Learning in NLP, we will present the recent advances in deep learning for NLP. The Section on Semantic Representation presents an overview in Semantic Representation Languages and Semantic Information Extraction in Natural Language Processing. Afterwards we will address new applications of NLP (exploring the use of the new advancements and semantic representation languages) to tasks that are relevant to the legal domain such as Legal Information Retrieval and Norm Extraction.

\section{Artificial Intelligence and Law}

The two traditional approaches to AI applied to Law, as a field, have been either Logic-based or Data-centric. The Logic-based approach has roots in the work of pioneers like \cite{Allen1957} but the first applications to real laws, like the formalization of the rules of the British National Act \cite{Sergot1986} were made in the eighties. The application of Data-centric approaches in the legal domain started also in the 1980s and is exemplified by the case-based reasoning HYPO system \cite{Ashley1991}. The goal of that system was to model reasoning on the basis of precedent, citing the past cases as justifications for legal conclusions.

\subsection{Logic-based approach}

\begin{figure}
\begin{center}
\includegraphics[width=150pt]{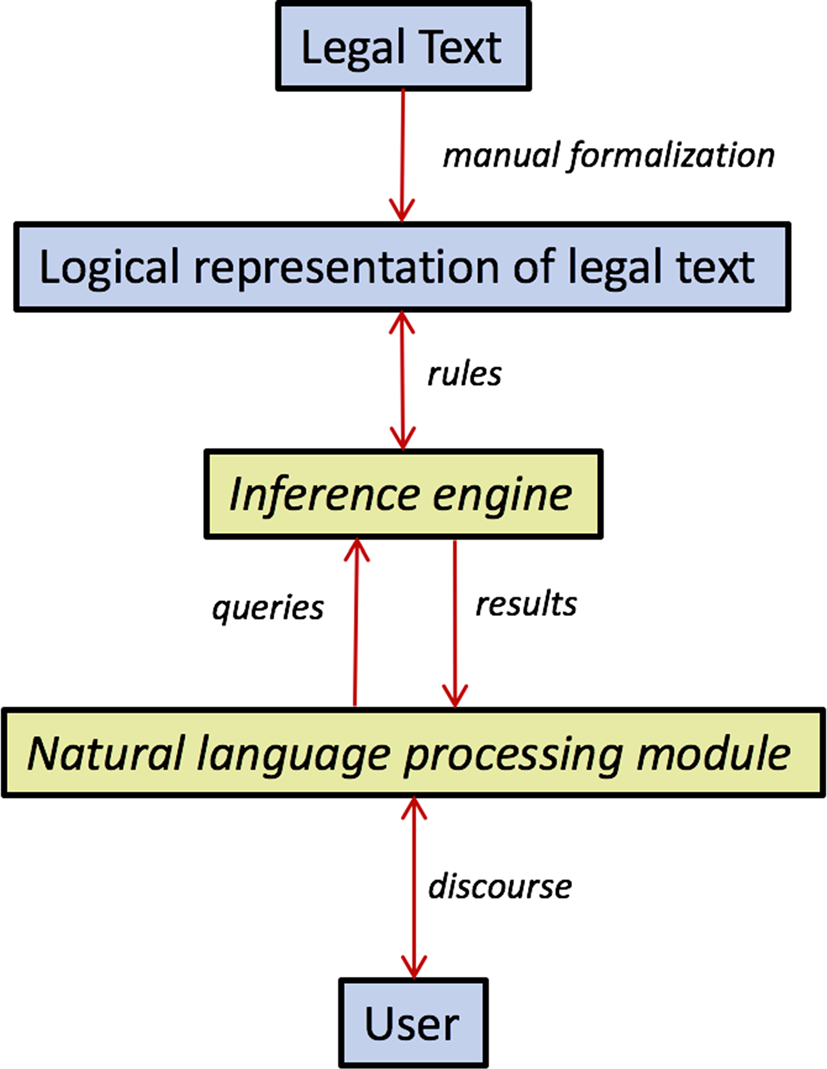}
\end{center}
\caption{Architecture of conversational Logic-based legal reasoning system. From \cite{Branting2017}}
\label{fig:Rule-based}
\end{figure}

Figure \ref{fig:Rule-based} shows the architecture of conversational Logic-based legal reasoning system. In principle, it would be possible to infer legal conclusions from a formalized version of the pertinent laws and of the case facts. But the ``natural language barrier" creates two major obstacles: the challenge of representing legal texts in the form of logical expressions and the difficulty of evaluating legal predicates from facts expressed in natural language \cite{Branting2017}.

\subsubsection{Converting between legal discourse and natural language}

A basic assumption of a logic-based approach is that it is possible to create a logical
formalization of a given legal text that is a faithful and authoritative expression of
the text’s meaning \cite{Branting2017}. This assumption can be divided in three parts:
\begin{enumerate}
    \item legal texts have a determinate logical structure;
    \item there is a logical formalism sufficiently expressive to support the full complexity of legal reasoning on these texts;
    \item the logical expression can achieve the same authoritative status as the text from
which it is derived.
\end{enumerate}

In fact, there is a many-to-many relationship between most legal text expressions and their logical representations. This implies that the logical formalization of the legal text, {\it unless enacted by the same legislative body}, will be just a one of many possible interpretations of the written legal text. This jeopardizes the first and third parts the assumption. Regarding the second part, a lot of work has been done in the last decades on logics specialized for the deontic, defeasible, and temporal characteristics of legal reasoning. But there is still no consensus regarding the most appropriate logic for a given purpose.

There is also the difficulty of the translation itself. It is a laborious and specialized task, needing deep knowledge of both legal writing and logic. Fortunately, recent advances in Natural Language Processing can conceivably be adapted to this task, as we will see below.

On the other extremity of the architecture, there is a second ``natural language barrier", between the abstract legal terms-of-art and ordinary discourse \cite{Branting2017}. Even if legal rules could be formalized with perfect fidelity, the terms in the rules are typically impossible for a layperson to interpret. Besides the difficulties with legal terminology, legal concepts are many times characterized by being deliberately vague or defeasible, in order to allow for social and technological changes. In fact, common interpretation of the legal discourse is are shared by legal practitioners within law schools and law courts and are inaccessible with only a literal reading of the legislative text \cite{Boella2019}.

Several ideas can be used to help  bridge this barrier, adjoining data-centric approaches and recent advances like case-based reasoning, translation, question answering and textual entailment.

\subsubsection{Some formal aspects of Legal norms}

In this section we discuss some formal aspects of Legal norms that a computerized system should be able to represent.

A legislative text can be viewed according to two different profiles \cite{Francesconi2014}:
\begin{itemize}
    \item a structural or formal profile, representing the traditional legislator habit of organizing legislative texts in chapters, articles, paragraphs, etc.;
    \item a semantic profile, representing a specific organization of legislative text substantial meaning; a possible description of it can be given in terms of normative provisions.
\end{itemize}

Provision types are organized into two main groups. On the first group there are Rules that can be either constitutive rules - introducing entities or regulative rules - expressing deontic concepts. On the second group there are Rules on Rules, which correspond to different kinds of amendments - content, temporal and extension (extends or reduces the cases in which the norm operates) amendments. 

\paragraph{Hohfeldian relations and Deontic Logic}

\begin{figure}
\begin{center}
\includegraphics[width=330pt]{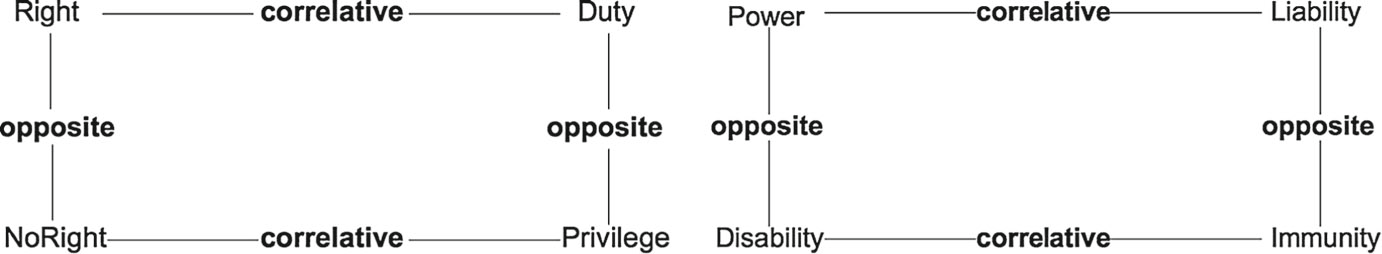}
\end{center}
\caption{Hohfeldian Relations. From \cite{Francesconi2014}}
\label{fig:hohfeldian-relations}
\end{figure}

A very important concept in legal theory are the Hohfeldian relations \cite{Hohfeld1913}. Hohfeld identified two quartets of fundamental relations. The first quartet involves deontic concepts in terms of correlative relations between Right and Duty, as well as No-right and Privilege, and opposite relations between Right and No-right, as well as Duty and Privilege. For example, A having a right towards B (e.g. receiving a payment) is equivalent to B having a duty towards A (making the payment). If A has a privilege (e.g. using a cellphone) towards B, it means that B cannot prevent A from doing so, so B has a No-right towards A (see Figure \ref{fig:hohfeldian-relations}).

The second quartet involves power relations, so that one can change right or power relations held by the other. This is expressed in terms of
correlative relations between Power and Liability, as well as Disability and Immunity, and opposite relations between Power and Disability, as well as Liability and Immunity. For example if A has a power towards B, this is equivalent to B having a liability (namely a subjection) with respect to A. Similarly, if A has an immunity with respect to a liability (subjection to a power of B), it means that B is disabled to limit A’s immunity.

To model Hohfeldian relations rules-based systems implement Deontic Logics, which are concerned with obligation, permission, and related concepts. An example is the description logic framework proposed by Francesconi \cite{Francesconi2014}.

\paragraph{Defeasible reasoning, hierarchies and Non-monotonic Logic}

Classical first order logic is monotonic because its inferences, being deductively valid, can never be "undone" by new information. That is, adding a new fact, consistent with the already existing facts, cannot invalidate any deductions already made. 

But many times legal text assumes and makes use of defeasible reasoning, where some conclusions can be invalidated by adding new facts. This type of reasoning allows for example for the definition of defaults, that is, rules that can be used unless overridden by an exception \cite{Polloc1987}.

Non-monotonic Logic can also represent priority relations between rules to resolve conflicts between different regulations within a normative system \cite{Prakken2015}.

\paragraph{The concept of time}

A normative system is never static. Laws are changed all the time and the same happens with legal interpretations. Legal obligations, permissions and other legal properties and relationships also exist in time, being triggered and terminated by the conditions contemplated by valid legal norms \cite{Prakken2015}. The following example, taken from \cite{Prakken2015}, illustrates well the complexity that needs to be modelled:

\begin{quote}
Consider, for instance, how a new regulation $R1$, issued at time $t_0$ may establish a new tax $a$ on transactions on derivatives, to be applied from time $t_1$. On the basis of $R1$ any such transaction happening after $t_1$ would generate an obligation to pay the tax, an obligation that will persist until the time $t_3$ when the tax is paid. Assume that at time $t_4$ a regulation $R2$ is issued which abolishes tax a and substitutes it with tax $b$. Then transactions taking place after $t_4$ will no longer generate obligations to pay $a$, but rather obligations to pay $b$... 
\end{quote}

There have been several proposals to incorporate the notion of time in Rule-based approaches. One proposal is to address temporal reasoning with norms using Event Calculus \cite{Hernandez1998}. Another is to extend defeasible logic to include temporal effects \cite{Governatori2007,Governatori2005,Rotolo2009}.

\subsection{Data-centric approach}

\begin{figure}
\begin{center}
\includegraphics[width=300pt]{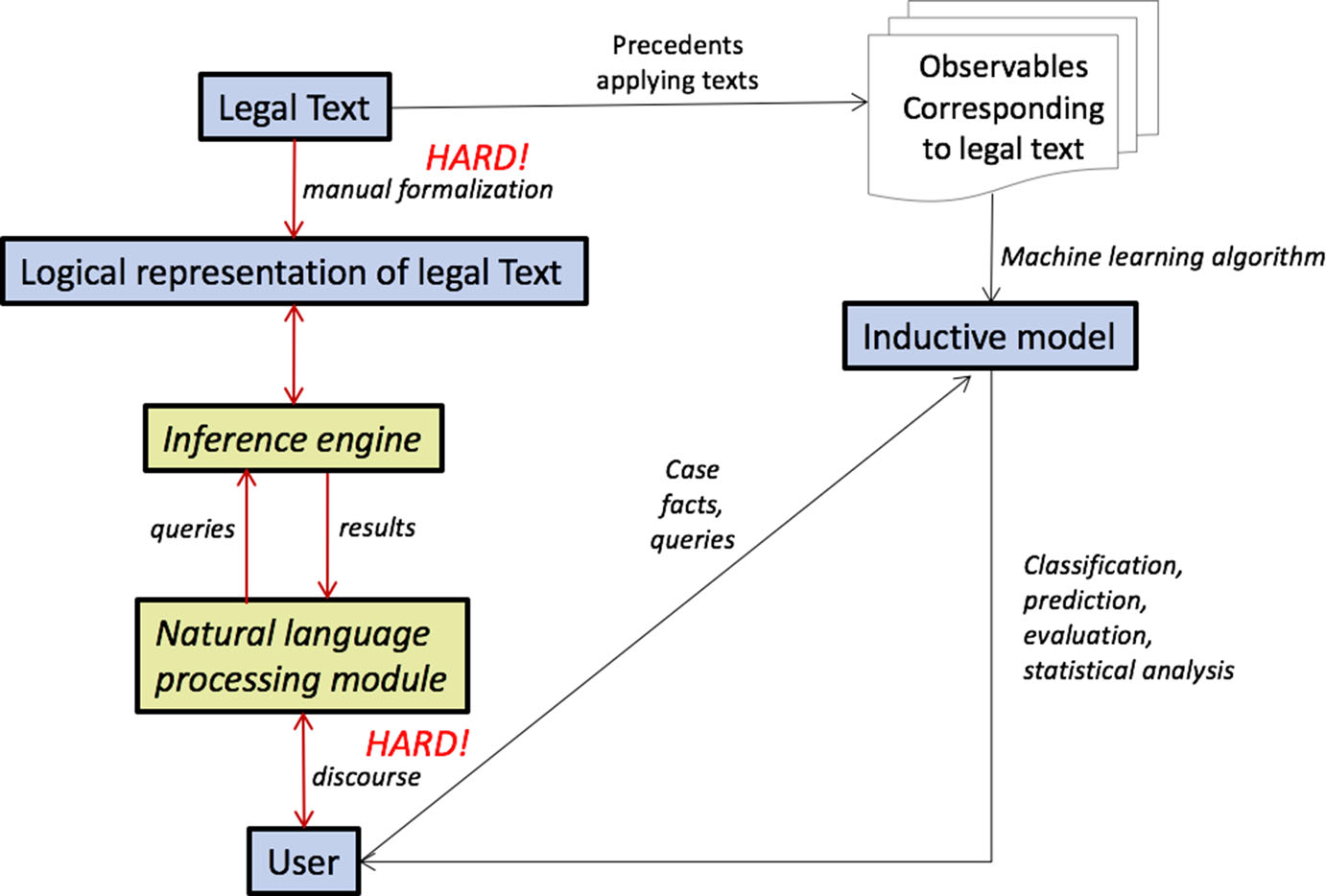}
\end{center}
\caption{Adding a data-centric component to the architecture of conversational Logic-based legal reasoning system. From \cite{Branting2017}}
\label{fig:Data-centric}
\end{figure}

Data-centric or data-driven approaches come with the benefits of automation and scalability,  greatly reducing the laborious human encoding of legal texts in logical formalisms. It can be used to replace particular components of a Legal Reasoning system or the whole system. As illustrated in figure \ref{fig:Data-centric}, the main premise is that by building a dataset of examples, for instance of a set of examples with legal text translated to a Logical Representation, one can then use machine learning techniques - such as deep neural networks - to inductively learn how to automatically perform the same task over new examples. 

However, there is a trade-off, because Data-centric approaches have in general lower transparency and explanatory capability \cite{Branting2017}. For instance, in the case of Deep Neural Networks, the learned model will be a very complex model with usually millions and even billions of parameters\footnote{As is the case of GPT-3 with 175 billion parameters} that are combined - often in a non-linear way - to determine a particular output. It is not easy for a human observer to understand why or how the model is producing a particular translation, prediction, or decision. For this reason, we argue that instead of replacing the whole system, a Data-centric approach may be better used as a tool to improve some components of a legal-reasoning system - for instance the component that translates legal text to its correspondent logical formalism - while integrating it with a traditional approach with a rule-based system (or inference engine) that will allow users to better understand the decisions being made by the system. 

\subsection{Representation Languages and Ontologies}

Currently, no single representation has been unanimously considered the de-facto standard for legal text. The closest to a legal standard format is the NormeInRete (NIR) format, used in Italy. The NIR format encodes the structural elements used to mark up the main partitions of legal texts, as well as their atomic parts (such as articles, paragraphs, and numbered items) and any non-structured text fragment. That format is implemented in XML.

Despite the lack of a standard representation for legal text, there has been an extensive work in the definition of representation languages and ontologies for the legal domain. There are also different level of approaches. Some authors look at the formalization of axiomatic ontologies using decription logis, while other focus on lightweight ontologies, i.e. a knowledge base storing low-level legal concepts, connected via low-level semantic relations, and related to linguistic patterns that denote legal concepts. 

The concepts that are most commonly used in legal ontologies are deontic operators - representing different types of prescriptions (obligation, prohibition, permission, exception), the active role and passive roles representing the addressee and beneficiary of prescriptions. In this section we briefly describe a non-exaustive list of some ontologies, markup languages, and representation languages defined for legal domains.

%The important questions are: (i) does the quasi-logical
%form capture in a convenient format the semantic informa-
%tion that will be needed by the information extraction sys-
%tem? and (ii) can the quasi-logical form be computed easily
%from the output of the statistical parser?
%\cite{McCarty2007}

\subsubsection{Legal Knowledge Interchange Format}

Legal Knowledge Interchange Format (LKIF) was an attempt to standardize the representation of legal knowledge in the semantic web. It was developed at the Estrella Project to be a knowledge representation formalism to help create an architecture for developing legal knowledge systems \cite{Hoeskstra2007}. 

LKIF was developed by asking different types of experts (citizens, legal professionals and legal scholars), to provide their top-20 set of legal concepts. The most relevant terms were identified, and together with concepts from other legal ontologies, they were used to form the main clusters for the development of the LKIF ontology. The different concepts were organized into a collection of ontology modules, each representing a different cluster of concepts: expression, norm, process, action, role, place, time and mereology. The top level modules help specify the context in which a legal fact, event or situation occurs by providing definitions of concepts such as location, time, parthood and change. Another set of modules, named the intentional level, include concepts and relations necessary for describing Actions undertaken by Agents with a given Role). These modules also specify concepts for describing mental states of these agents, such as their Intentions and Beliefs. The Legal level is concerned with defining concepts specific to the legal domain: legal agents (e.g. natural person vs legislative body), legal actions, rights and powers, legal roles and norms. Similarly to other legal ontologies, LKIF models norms as being deontic qualifications of situations, allowing to represent that a certain situation is Obliged, Allowed, Prohibited or Disallowed.    

%see https://books.google.pt/books/about/Legal_Ontology_Engineering.html?id=JBR2zq-voVQC&printsec=frontcover&source=kp_read_button&redir_esc=y#v=onepage&q=LKIF&f=false  for more information

\subsubsection{Legal-URN}

Legal-URN (Legal User Requirements Notation) \cite{Ghanavati2013} is not an ontology, but a modelling framework based on a Requirements Engineering approach to legal compliance, modeling legal norms using notation from goal and business process management. What makes Legal-URN interesting is the combination of an Hohfeldian model used for the specification of norms with a Legal Goal-Oriented Requirement Language (Legal-GRL), which tries to capture the objectives and requirements of the legislation. Table \ref{table:hohfeldian-Legal-URN} depicts an example of the Hohfeldian model used by Legal-URN.

\begin{table}
\footnotesize
\centering
\begin{tabular}{|l l|}
\hline
\textbf{Hohfeldian Model} & \textbf{Example} \\
\hline
Section & Organizational Requirements\\
Article\# & DIRECTIVE 2004/39/EC - 13\\
Subject & An Investment Firm\\
Modal verb & Shall\\
Clause & arrange for records to be kept of all services, [...]\\
Precondition & -\\
Exception & -\\
XRef & -\\
PostCondition1 & Be sufficient to enable monitor compliance\\
PostCondition2 & Ascertain that [...] with obligation [...] to clients\\
\hline
\end{tabular}
\caption{Hohfeldian model example used in Legal-URN. From \cite{Ghanavati2013}.}
\label{table:hohfeldian-Legal-URN}
\end{table}

\subsubsection{LegalRuleML}
LegalRuleML \cite{Athan2015} is a markup language designed to be a standard for legal knowledge representation, harmonizing different types of legal texts. It provides a rich set of concepts useful for a formal representation of legal-texts. Of interesting notice is LegalRuleML's capability to support multiple legal interpretations of a legal rule, by allowing multiple semantic annotations.

LegalRuleML is also able to handling defeasability - detecting a resolving conflicts between legal rules - by providing the element Overrides, which defines a relation of superiority where a legal statement l1 overrides another legal statement l2, and the elements DefeasibleStrength and Defeater to specify what should be considered to hold or not hold in the absence of information. Table \ref{table:LegalRuleML-elements} illustrates some of the elements that can be used to annotate legal rules in LegalRuleML. 

\begin{table}
\footnotesize
\centering
\begin{tabular}{|p{1.5cm} p{9.5cm}|}
\hline
\textbf{Element} & \textbf{Meaning} \\
\hline
Agent & an entity that acts or has the capability to act\\
Alternatives & a mutually exclusive collection of one or more Legal Norms\\
Authority & a person or organization with the power to create, endorse, or enforce Legal Norms\\
Bearer & a role in a Deontic Specification to which the Deontic Specification is primarily directed\\
Compliance & an indication that an Obligation has been fulfilled or a Prohibition has not been violated\\
Defeater & an indication that, in the absence of information to the contrary and where the antecedent of a Legal Rule holds, the opposite of the conclusion of the Legal Rule does not hold\\
Obligation & a Deontic Specification for a state, an act, or a course of action to which a Bearer is legally bound, and which, if not achieved or performed, results in a Violation \\
\hline
\end{tabular}
\caption{Example of elements from LegalRuleML and their meaning. From \protect\footnotemark.}
\label{table:LegalRuleML-elements}
\end{table}

\footnotetext{http://docs.oasis-open.org/legalruleml/legalruleml-core-spec/v1.0/legalruleml-core-spec-v1.0.html}

\subsubsection{European Legal Taxonomy Syllabus}

\begin{figure}
\begin{center}
\includegraphics[width=345pt]{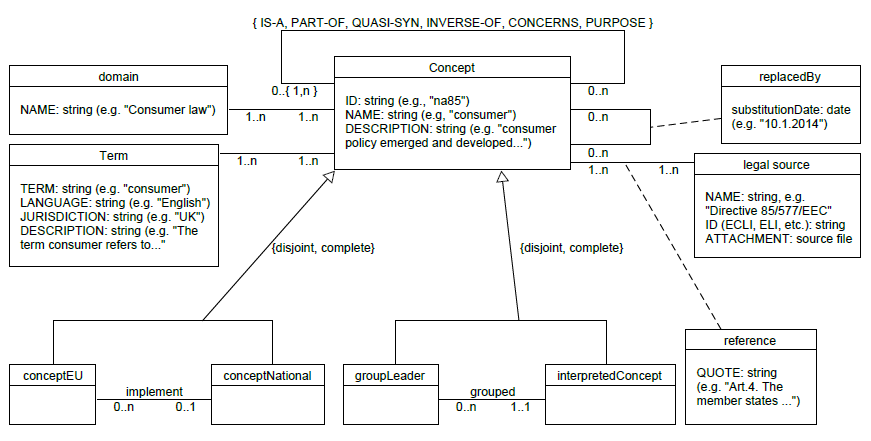}
\end{center}
\caption{UML Representation of ELTS Ontology Schema. From \cite{Ajani2016}}
\label{fig:ELTS_Schema}
\end{figure}

The European Legal Taxonomy Syllabus (ELTS) is a a specialist multilevel multilingual ontology \cite{Ajani2016}. LTS is a a lightweight ontology and was designed with the goal of modeling the translation and implementation of European directives into national law using terms that are defined within national legal systems. To that end, one of the key characteristics of the ELTS Ontology is the modeling of terms and concepts as independent entities that can have many to many relationships (i.e. a term can correspond to multiple concepts, a concept might have multiple terms). Figure \ref{fig:ELTS_Schema} depicts the UML representation of the ELTS Ontology Schema. In addition to traditional ontological relations such as ``is-a'', ``part-of'', the ELTS Ontology also models more specific legal relations such as legal ``purpose'', and meta-relations such as the implementation relation - representing how a concept is transposed into a national legal system.

\subsubsection{Ontomedia}

The Ontomedia semantic platform \cite{Fernandez-Casanovas2011} was created with the goal of establishing an ontology based of non-expert generated content in the domain of consumer mediation. In this platform – which functions like a community where citizens, administrations, institutions and professionals can intervene – the users (citizens) provide input texts describing their conflicts or asking for institutional assistance and are redirected to the relevant information available online regarding the exposed problem or the suitable state agency.

\begin{figure}
\begin{center}
\includegraphics[scale=0.65]{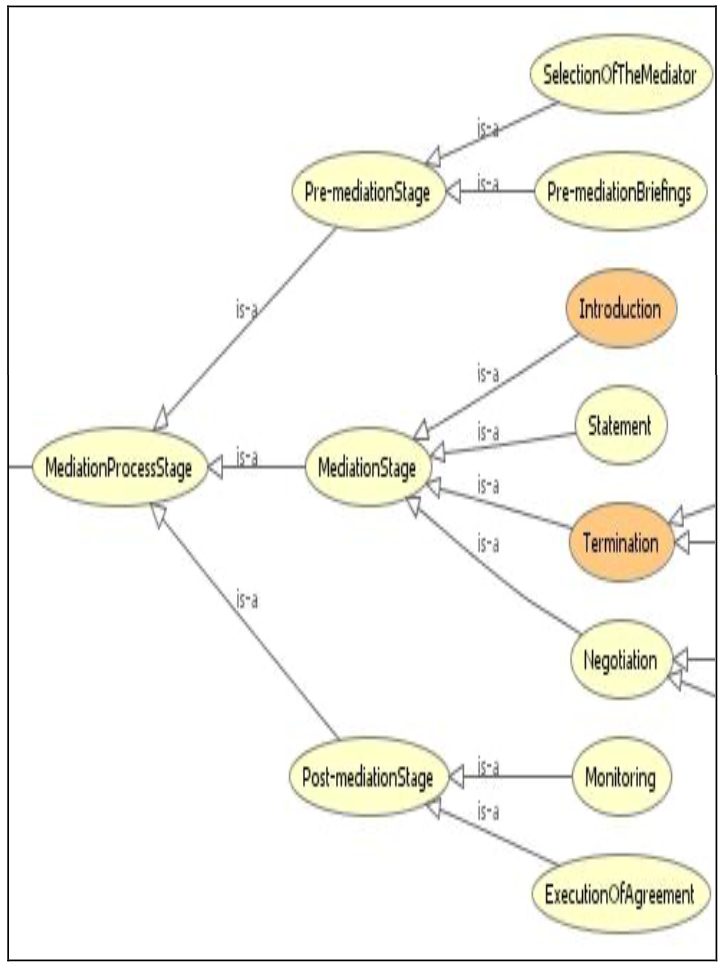}
\end{center}
\caption{Fragment of the Mediation-Core Ontology from \cite{Fernandez-Casanovas2011}}
\label{MC_ontology}
\end{figure}

To make this possible, the Mediation-Core Ontology (MCO) and Consumer Mediation Ontology (CMO) ontological structures are used as an ontological base. These ontologies allow the mapping of the user’s set of linguistic units – which are used to represent the query or expose the user’s conflict – into an institutional representation of the domain. Whereas the CMO ontology focuses on the particularities of the mediation in the consumer domain (with classes denoting the parties involved in the conflict, the regulation applicable, the geographic area and the type of conflict), the MCO ontology, pictured in Fig. \ref{MC_ontology}, is more focused towards the mediation process and its' stages.

\begin{figure}
\begin{center}
\includegraphics[scale=0.6]{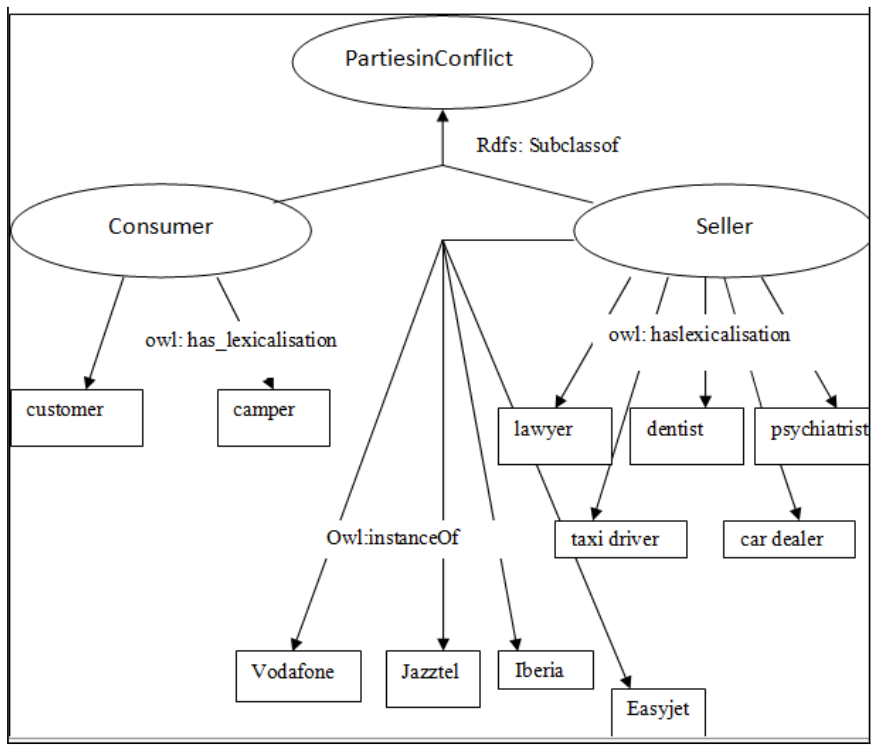}
\end{center}
\caption{Integration of consumer terminology into the CMO ontology from \cite{Fernandez-Casanovas2011}}
\label{CM_ontology}
\end{figure}

Since recent studies in the field are more focused towards generating ontologies derived from legal texts, and these contain a formal terminology – unlike most consumers’ complaints and questions that are formulated with day-to-day terminology – the authors enriched the CMO with non-formal terminologies used by the consumers in their corpus, like shown in Fig. \ref{CM_ontology}. This was done by extracting terms based on their morphological tag, in particular nouns, and is what allows identifying the domain of the question or complaint. For example, the terms flight, plane and airport are related to the Passenger Air Transport topic whereas the words Internet, portability and Vodafone are related to the Telephone and Internet topic.

\subsection{Current AI-based legal tools}

Our research revealed almost no AI-based legal tools for the common public. There are online legal information systems, like the one developed by the University of Delhi for the Indian legal context \cite{Bhardwaj2016}, but they do not use AI technologies.

\subsubsection{Menslegis/Eunomos}

Menslegis is a commercial service for compliance distributed by Nomotika s.r.l., a spinoff of the University of Turin, in Italia, incorporating the Eunomos system \cite{Boella2019}. As described by one of its creators, it is a document and knowledge management system, with a web interface for legal researchers and practitioners to manage knowledge about legislation and legal concepts. The software improves access to legislation and understanding of norms. It enables users to search and view relevant legislation from various sources from an internal database (containing more than 70,000 Italian national laws), where legislation is classified and enriched in structure. To deal with continuous updates in legislation, it allows legislative amendments to be automatically recognized and retrieved. Furthermore, The system offers users access to a database of prescriptions (duties and prohibitions), annotated with explanations in natural language, indexed according to the roles involved in the norm, and connected with relevant parts of legislation and case law. It also offers an ontology of legal concepts that are relevant for different domains in compliance. Finally, terms within the legislation are linked to the concept descriptions in the ontology.

\begin{figure}
\begin{center}
\includegraphics[width=250pt]{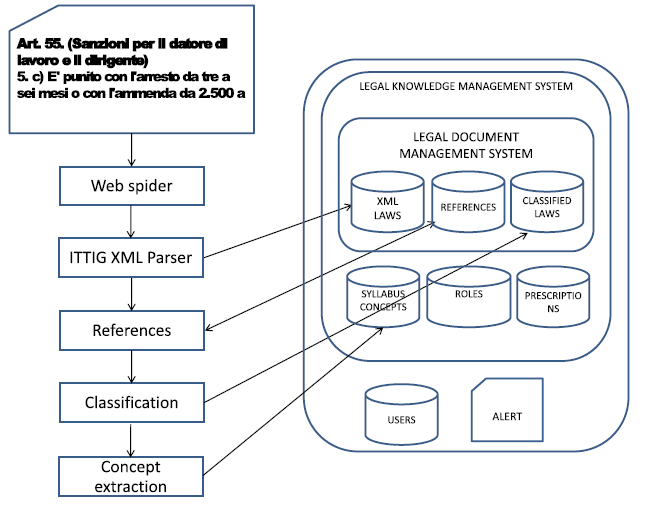}
\end{center}
\caption{Architecture of Menslegis/Eunomos system. From \cite{Boella2019}}
\label{fig:Menslegis}
\end{figure}

The architecture of the system comprises three levels:
\begin{enumerate}
    \item The legal document management system, composed of a database of norms in legislative XML, a database of references between laws—using their unique processable identifier called URN (see Sect. 4.4), and a database classifying legislation or single articles within legislation in accordance with the Eurovoc Thesaurus;
    \item The legal knowledge management system is composed of a database of concepts and the relations connecting them, together with the terms associated with the concepts. This level also includes a database of prescriptions (obligations);
    \item The external tier is composed of a database of user profiles with login details and information about users’ domains of interest. This tier also includes a functionality for dispatching alerts to users about updates in legislation of interest to them.
\end{enumerate}

The Eunomos system uses the ELTS ontology with two extensions. Firstly, it incorporates the Eurovoc Thesaurus—a multilingual, multidisciplinary thesaurus with about 7000 categories covering the activities of the EU. This is used as an independent means to classify documents. Secondly, the ELTS ontology framework was extended to model not only definitions of legal terms but also to describe prescriptions, taking inspiration from the way compliance officers in financial institutions extract norms for regulatory compliance purposes.

\subsection{Traditional Approaches to Legal Information Retrieval}

Legal Information Retrieval (LIR) is a specific type of Information Retrieval and, therefore, requires different approaches to the way the text is searched. Usually, legal documents are written in a very formal language but search queries written by regular citizens tend to be in a more informal language. This type of imbalance creates a mismatch of vocabulary that damages search if not attended to. Traditional LIR systems are mostly based on two techniques --- Boolean Search and Manual Classification.

In Boolean Search, documents are scanned through in order to find an existing match in the search terms, according to the logical conditions imposed. A user may specify terms such as specific words or judgments by a specific court. These terms are combined with logic symbols such as AND, OR, NOT in order to provide a search on the content of the texts. They are widely implemented by services such as Westlaw\footnote{https://legal.thomsonreuters.com/en/products/westlaw}, LexisNexis\footnote{https://www.lexisnexis.com/en-us/gateway.page}, and FindLaw\footnote{https://www.findlaw.com/}, which are American legislation search services. This kind of search performs poorly when literal term matching is done for query processing, due to synonymy and ambivalence of words.

Manual classification is a technique that is used to overcome the limitations of Boolean searches. This technique relies on classifying case laws and statutes into computer structures. This structuring is done according to the way a legal professional would organize them and it also attempts to link the texts based on their subject or area. It allows the texts to be classified and it makes it easier to extract knowledge for a search. The reason why this technique may be considered unsustainable is because there is an increasing amount of legal texts and not enough legal professionals or time to classify them.

When we look back from Legal Information Retrieval into the broad spectrum of Information Retrieval (IR), other options appear. These options do not have a focus on the legal subject but rather on generic text documents. In spite of the particularities of legal documents, such as the connections between different documents, for instance, we can also look at legal articles as text documents --- therefore expanding the field of research. When we do this, we are met with several other studied alternatives.

The Okapi BM25 \cite{okapibm25}, or rather just BM25 (BM stands for best matching) is one of those alternatives and it is widely used as an IR ranking algorithm. This algorithm is still used today by search engines to determine the relevance of entries to the searched query, and it is still considered to be a state-of-the-art document retrieval function, along with TF-IDF (Term frequency - inverse document frequency), on which it relies, as we will see.

BM25 is a bag-of-words retrieval algorithm, which is defined by the representation of text as a set (or \textit{bag}) of words while disregarding their syntax or context. The most popularized version of BM25, introduced in TREC 1994 (a conference on text retrieval) is the following:

Given a query \(Q\), containing keywords \(q_1,q_2,...,q_n\), the BM25 score of a document \(D\) is defined as:
\[score(D,Q) = \sum_{i=1}^{n}  IDF(q_i) \cdot \frac{f(q_i, D) \cdot (k_1 + 1)}{f(q_i, D) + k_1 \cdot (1 - b + b \cdot \frac{|D|}{avgdl})}\]

\noindent where \(f(q_i, D)\) is the term frequency of \(q_i\) in the document \(D\), \(k_1\) and \(b\) are optimization parameters, \(|D|\) is the length of the document \(D\) in words and \(avgdl\) is the average document length in the set of documents. \(IDF(q_i)\) is the inverse document frequency of the term \(q_i\). It is used as a weight function and it is defined as:

\[IDF(q_i) = log\frac{N - n(q_i) + 0.5}{n(q_i) + 0.5}\]

\noindent where \(N\) is the cardinality of the set of documents and \(n(q_i)\) is the number of documents that carry the term \(q_i\).

In a paper \cite{colieebm25bert} released for the COLIEE workshop in 2019, this algorithm was used to retrieve legal information based on query search. In the third task (first task of statute law) of the competition, they used the BM25 function to derive the score of each document based on a searched query. It delivered promising but not outstanding results compared to other IR techniques.

\subsection{Legal Prediction and Case Law Analysis}

Documents detailing precedent court decisions, case law, are key in common law legal systems adopted by the USA, UK, etc. However, case law also plays a role in civil law countries by guiding the permissive interpretations. This role is discussed in \cite{ashley2004case} where several civil law and common law jurisdictions are reviewed. Nowadays there are large publicly available collections of precedents in digital form. This abundance of (unstructured) information opens the door for the usage of data-driven approaches to automatically or semi-automatically analyze legal data.

One application of these data-driven approaches is to optimize the evidence-based search done by lawyers. Without proper searching tools and filters, surround oneself with as much relevant information from case law documents as possible takes long.
The work \cite{barros2018case} tackles this problem by building a classifier to infer the most important features for document classification. 
The work use case law documents from a national court and documents were represented using TF-IDF. A Bayesian Network was trained, achieving 90\% accuracy. 

Another way of using precedent cases is for judicial decision-making. This task is called Legal Judgement Prediction (LJP) and consists of automatically predict the outcome of a case by solely relying on the case document.
Prior work that choose to follow the machine learning route tend to use more classical methods \cite{medvedeva2020using, visentin2019predicting, strickson2020legal}. The most popular method is the Support Vector Machine (SVM), which is used in \cite{aletras2016predicting}. This work caught the attention of the media at the time it was published. The authors used a small dataset of 250 cases from the European Court of Human Rights (ECHR) and used a SVM to classify them as having a violation case or not. The model achieved 79\% accuracy.

Another impressive result was obtained in \cite{DBLP:journals/corr/abs-1708-01681} where an SVM was used to predict the law area and the decision of cases judged by the French Supreme Court. The model achieved 96\% f1 score in predicting a case ruling, 90\% f1 score in predicting the law area of a case, and 75.9\% f1 score in estimating the time span when a ruling has been issued.

Although less popular, there is some prior work that use deep learning for LJP. One example is the work \cite{chalkidis2019neural} where authors use 11.5k cases from ECHR’s public database\footnote{https://www.echr.coe.int/Pages/home.aspx?p=caselaw/HUDOC\&c=} to test various neural networks (e.g. BERT) for binary violation classification, i.e. classify a case law document has having or not a violation. The neural models considered outperform previous feature-based models.

The previously mentioned works require documents to be represented in a way the models understand. As previously said, some works use $N$-grams and TF-IDF. Bag of words is also a popular choice. There are also projects that extract a topic from the document and use it as a feature. In \cite{grajzl2020machine}, authors create a corpus of 52,949 reports of cases heard in England’s high courts before 1765 and use a 100-topic structural topic model to attribute topics to these documents. 

Using machine learning has the great disadvantage of offering little to no explanation of how the result was obtained. In \cite{branting2020scalable} an attention based network is used both for prediction and highlighting the most crucial text considered by the network when classifying the document. However, the results obtained did not improve human decision speed or accuracy as some participants find it difficult to relate the highlighted text with the task they were given. Moreover, law evolves through time and systems trained on old cases will have their accuracy decreased. 
For these reasons, there is work that deviates from the machine learning approach and follows an argumentative-approach by using rules. In \cite{branting2020scalable}, authors use the ANGELIC methodology (detailed in the paper) to build a system capable of displaying it's arguments for classifying a document in the same manner as the work \cite{aletras2016predicting} did with an SVM. Evaluation was not extensive but results suggest higher accuracy than \cite{aletras2016predicting}.

\section{Advances of Deep Learning in NLP}
In the last few years, researchers have seen a major breakthrough in the application of deep learning models and techniques to the area of Natural Language Processing, due to the appearance of the transformer-based models. This section will present an overview of these progresses by describing the evolution of encoder-decoder models used in NLP, starting from sequence-to-sequence models based on LSTMs and ending in the most recent transformer-based approaches.  

\subsection{Encoder-Decoder and Sequence to Sequence Models}
Although often wrongly used interchangeably, the terms Encoder-Decoder and Sequence to Sequence represent two different model concepts. An Encoder-Decoder model is composed by two neural networks (the encoder and the decoder) and by an internal fixed-sized vector that represents the latent space. 

The encoder processes an input which could be a a word, a sentence or a document, and generates a contextualized vector representation of the given input. When trained effectively, this vector representation produced by the encoder will capture the most relevant features of the input. The decoder then uses this internal vector representation as input to generate an appropriate output for a specific task.  

A Sequence to sequence network, is a specific type of encoder-decoder that is able to deal with input and output sequences of arbitrary length.
Proposed in \cite{Sutskever2014}, it learns a transformation from one representation to another. This architecture is widely used in NLP in order to transform a sequence of words into a numerical representation, since it's capable of generating arbitrary length contextual vectors as an output sequence.  It has been applied to a wide range of tasks like machine translation \cite{Sutskever2014} and image caption generation \cite{venugopalan2015sequence}, reaching state of the art results.\par
\begin{figure}
\includegraphics[width=0.8\textwidth]{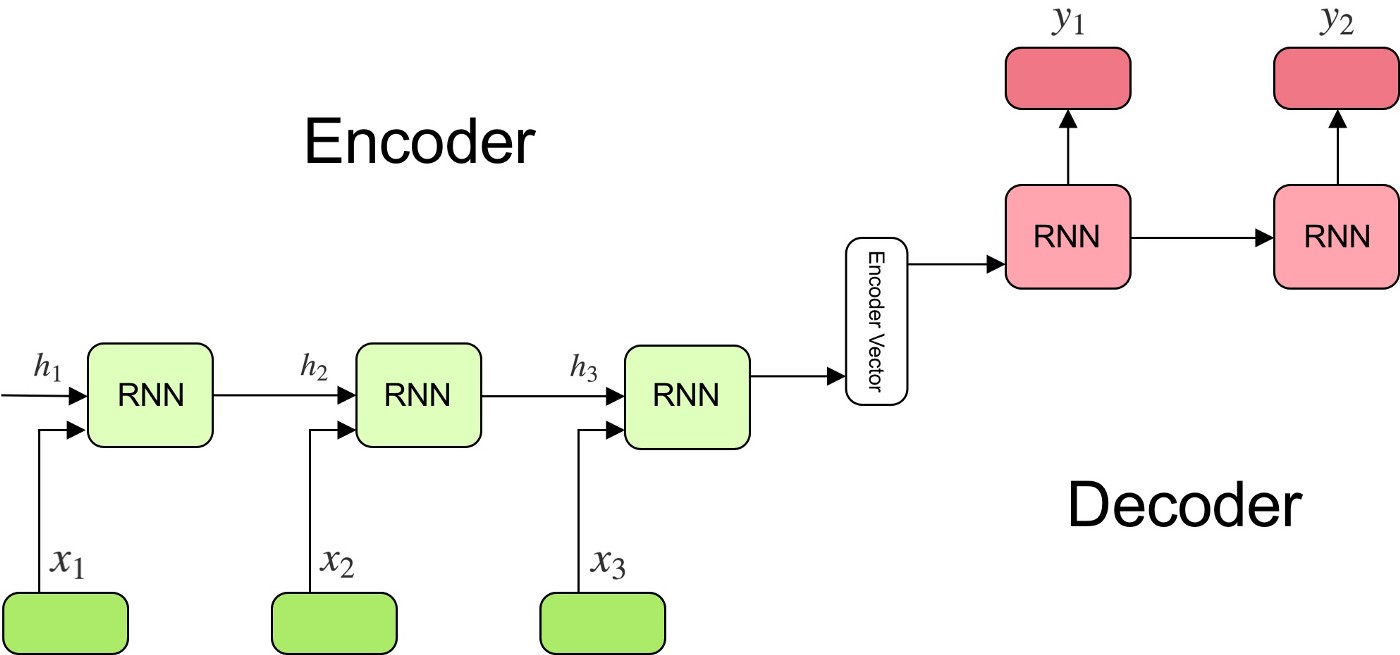}
\centering
\caption{Sequence to sequence model architecture from Simeon Kostadinov\protect\footnotemark  that generates a sequence output from a sequence input. The input embedding is passed onto the decoder through the encoder network's final hidden state}
\centering
\label{fig:seq2seq}
\end{figure}
\footnotetext{https://towardsdatascience.com/understanding-encoder-decoder-sequence-to-sequence-model-679e04af4346}
As illustrated in Figure \ref{fig:seq2seq}, these networks are composed by an encoder and a decoder.

Encoder-Decoder Networks must capture the complete sequence of information in a single vector, which leads to a problem when for example, encoding long-range dependencies or when we need to hold on to information that sits in the beginning of the sequence. \par
This architecture is an alternative to traditional statistical machine translation and it's been widely explored since its appearance in 2014 because it is more sensitive to word order, syntax and meaning of the source sentence, allowing users to capture more contextual information when creating a language model.

\subsection{Recurrent Neural Networks}
A Recurrent Neural Network (RNN) is a type of neural network that excels at processing sequences of data with its connective structure that allows information to be passed from the previous input to the current. RNNs can also be seen as multiple feed-forward neural networks that flow information from one network to the other, when in truth they are simply one network in which the cells (nodes) loop over themselves for every input element that it receives. 

These networks are commonly used for tasks that require pattern recognition in which elements in a sequence depend on each other's value.

\begin{figure}
\begin{center}
\includegraphics[width=150pt]{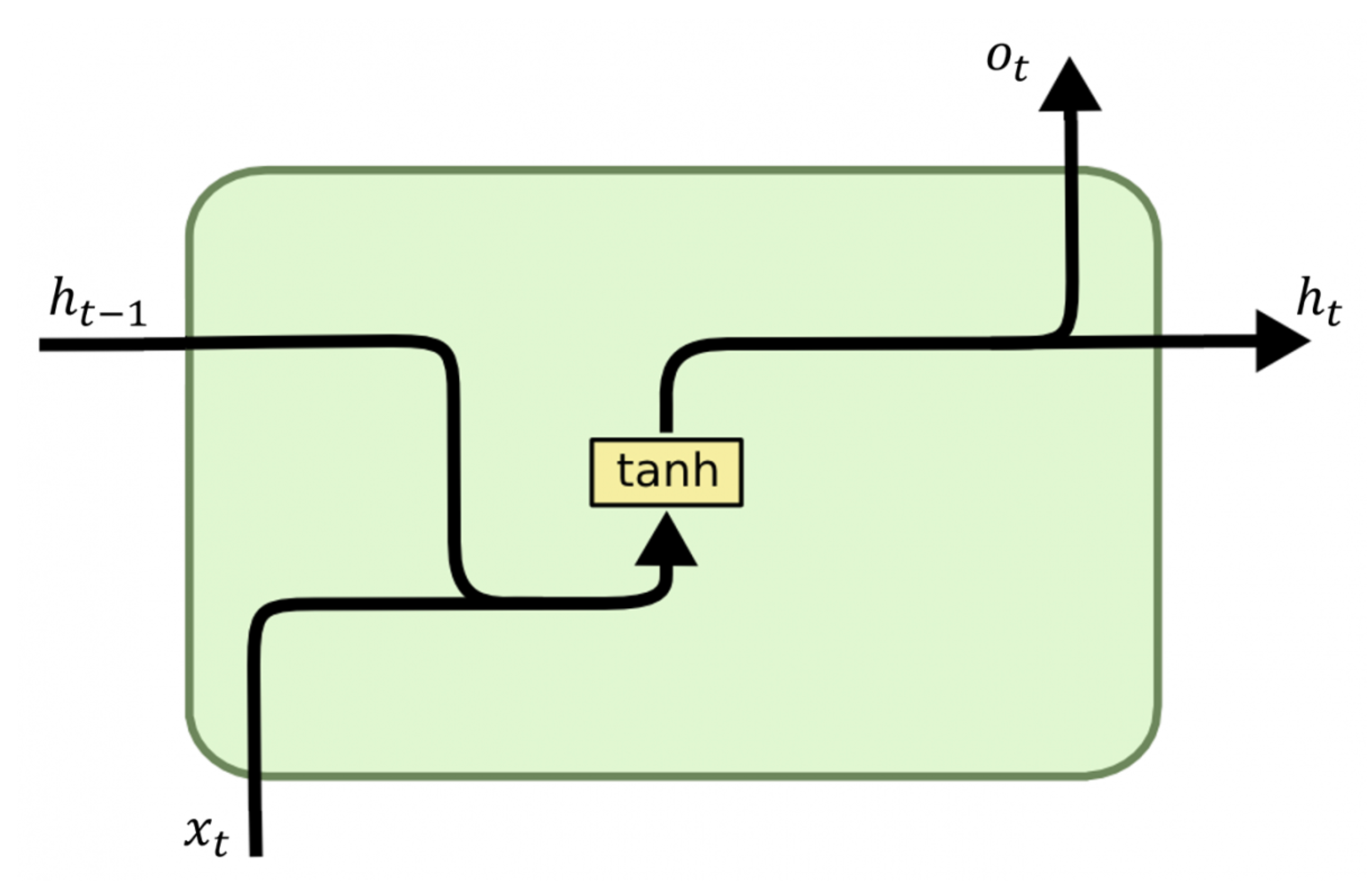}
\end{center}
\caption{RNN cell.\protect\footnotemark}
\label{rnn}
\end{figure}
\footnotetext{From http://dprogrammer.org/rnn-lstm-gru}

RNNs use a hidden state vector to pass information from the previous state to the next. The hidden state vector, \(h_t\), depends on the previous hidden state, \(h_{t-1}\), and the current input, \(x_t\), and therefore holds information about the previous state. It is defined as:

\[h_t = f(W_{hh} \cdot h_{t-1} + W_{hx} \cdot x_t + b_h)\]

\noindent where \(f\) is a non-linear activation function that is usually a hyperbolic tangent function, tanh, or a sigmoid function, \(\sigma\).

The output vector (prediction), \(o_t\), depends on the hidden state vector, \(h_t\).

\[o_t = g(W_{oh} \cdot h_t + b_o)\]

\noindent where \(g\) is another activation function, normally \(softmax\).

The different weight matrices, \(W\), are initialised randomly and adjusted using the error from the loss function. The added \(b\) in each formula are biases.

In the training of a RNN, the weights and biases are adjusted through a process called Back-Propagation Through Time (BPTT) that is heavily based on the standard back-propagation method, in which the weight matrices are updated according to the loss function --- the error between the predicted and actual observations.

In a summarized version of the algorithm, for each time-step we calculate the error. Those errors are then accumulated through the time-steps and a gradient of the loss function is calculated for each weight matrix (a partial derivative of the loss function with respect to the weight matrix). The gradient of each weight matrix is used to update the values of the matrix, according to a learning rate. When the error between prediction and observation is satisfied, we stop.

However useful these networks may be for some tasks, they pose an issue for others in which the values in a sequence have long-term dependencies (values far apart in a sequence depend on each other). This issues arises when the back-propagated errors start to diminish along the layers (tend to zero) or increase in an exponential way, which inhibits the weights from updating --- the so-called Vanishing/Exploding Gradient Problem. This, in turn, renders the training useless and long-term dependencies get ignored or over-estimated in the process of prediction. For instance, in next-word prediction, a word is predicted based on all of the previous words in a sentence, e.g., in the sentence, \textit{Yesterday, I walked my dog}, when trying to predict the word \textit{dog}, a network would have to consider all of the words that came before it. Nonetheless, the closest words tend to be more relevant for the prediction but that is not always the case. In some cases, words that are far apart in a sentence have a high dependency, such as the words \textit{dog} and \textit{bones} in the sentence \textit{My dog is very smart and he likes bones}. These long-term dependencies are not easily learnt by RNNs. 

This example was demonstrative only and does not represent the real challenge of long-term dependencies. In a real setup, long-term dependencies can be between thousands of elements in a sequence.

\subsection{Long Short-Term Memory}
\begin{figure}
\begin{center}
\includegraphics[width=200pt]{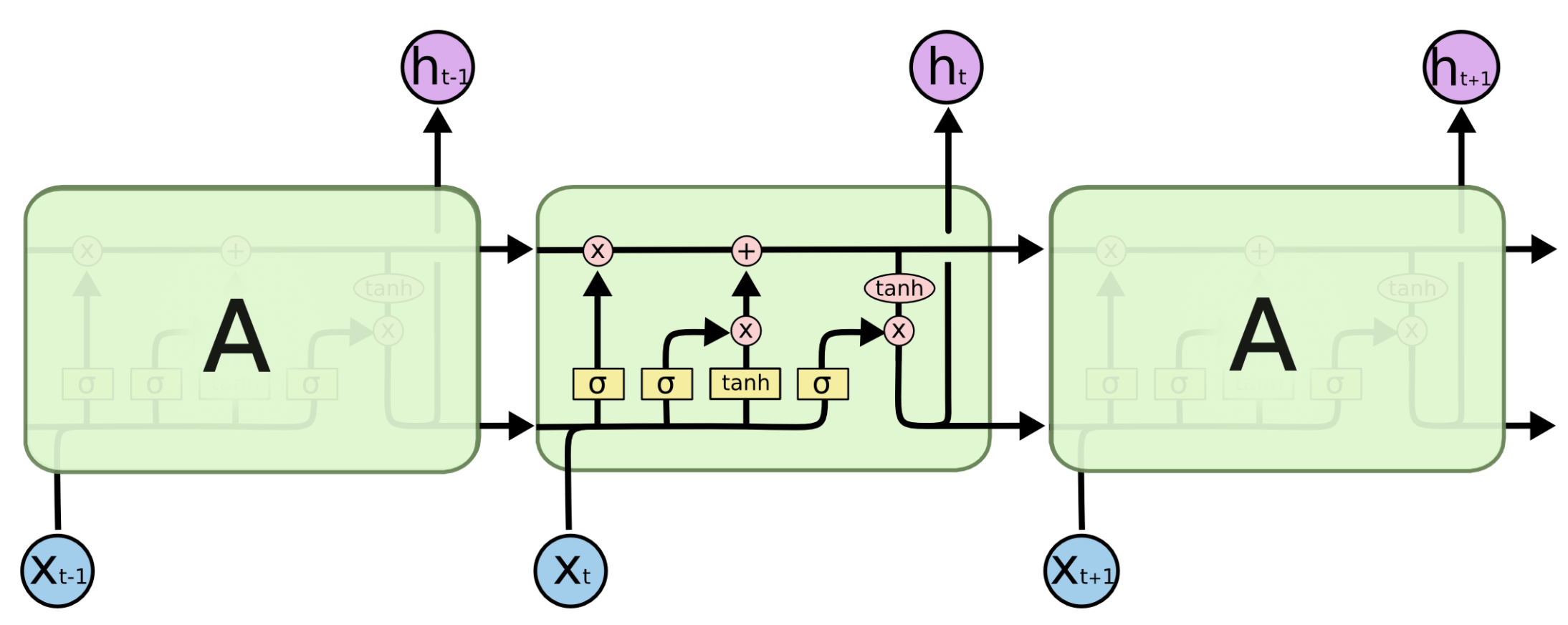}
\end{center}
\caption{LSTM Structure.\protect\footnotemark}
\label{lstm}
\end{figure}

Long Short-Term Memory (LSTM) \cite{hochreiter1997long} is a special type of a RNN that employs a gating mechanism when flowing information between time-steps. Apart from the hidden state vector present in RNN cells, the LSTM cell is characterized by having an additional cell state vector that, at each time-step, can be read, written or reset --- depending on the gate. (see Figure \ref{lstm}).

\footnotetext{From http://colah.github.io/posts/2015-08-Understanding-LSTMs/}

LSTMs overcome the Vanishing/Exploding Gradient Problem by allowing the errors to back-propagate unchanged, resulting in the update of long-term weights that need to exist in order to learn long-term dependencies.

\begin{figure}
\begin{center}
\includegraphics[width=150pt]{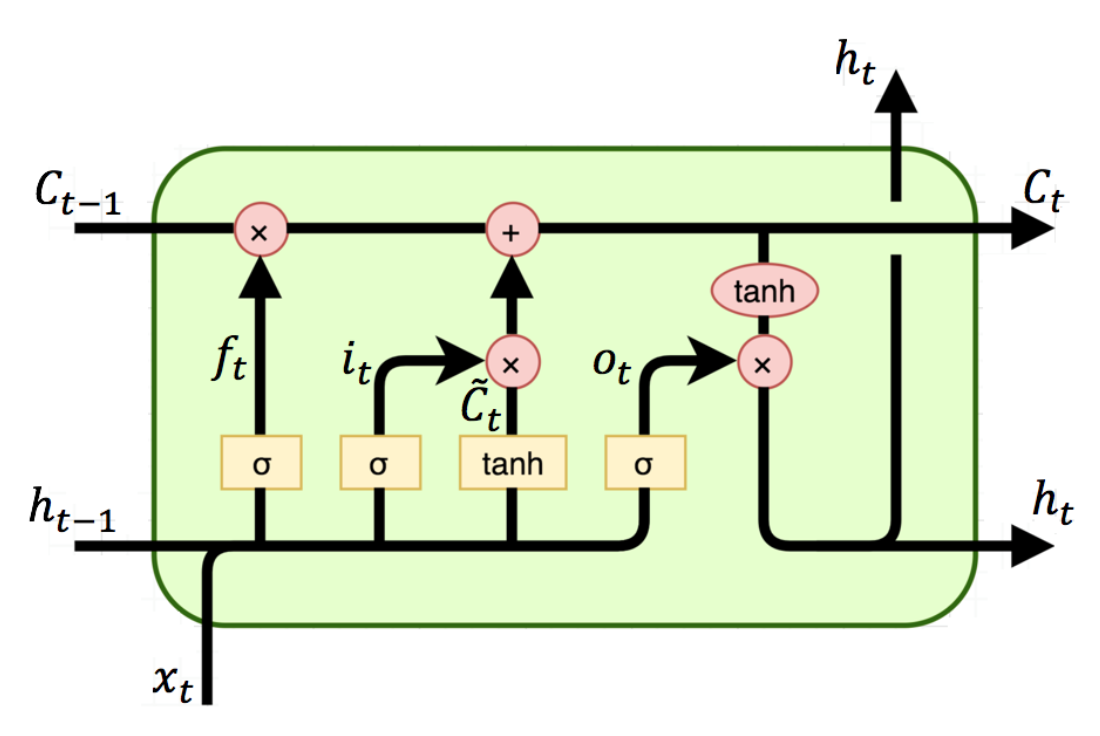}
\end{center}
\caption{LSTM cell.\protect\footnotemark}
\label{lstm_overview}
\end{figure}

LSTM networks work pretty similarly to standard RNNs with the exception of the structure of the cells. In a LSTM cell, there are three major gates --- the forget gate, the input gate and the output gate (See Figure \ref{lstm_overview}).

The gates are merely vectors, like the others in the structure, but they are called gates given that they control the amount of information that flows between cells at different time-steps.

\footnotetext{From http://dprogrammer.org/rnn-lstm-gru}

The forget gate, \(f_t\), is defined as:

\[f_t = \sigma (W_f \cdot [h_{t-1},x_t] + b_f)\]

The input gate, \(i_t\), is given by:

\[i_t = \sigma (W_i \cdot [h_{t-1},x_t] + b_i)\]

The new cell state candidate vector, \(\tilde{C}_t\), is given by:

\[\tilde{C}_t = \text{tanh} (W_C \cdot [h_{t-1},x_t] + b_C)\]

With these values, we are then able to calculate the cell state, \(C_t\).

\[C_t = f_t \cdot C_{t-1} + i_t \cdot \tilde{C}_t \]

The forget gate, \(f_t\), defines how much of the previous cell state, \(C_{t-1}\), goes into the current one, and the input gate, \(i_t\), determines how much the cell state candidate, \(\tilde{C}_t\), should contribute to the cell state.

Ultimately, we have the output gate, \(o_t\) which is responsible for controlling what parts of the cell state get to be transmitted to the next cell.

\[o_t = \sigma (W_o \cdot [h_{t-1}, x_t] + b_o)\]

This output vector is then used to calculate the hidden state vector, \(h_t\), which is transmitted to the next cell.

\[h_t = o_t \cdot \text{tanh} (C_t)\]

\(W_x\) and \(b_x\) with  \(x \in \{i,f,o,C\}\), are the weight matrix and bias vector of the input, forget, output and cell state vectors, respectively. The weights and biases will get adjusted during training. \(h_{t-1}\) is the hidden state of the previous cell and \(x_t\) is the input vector of the current cell.

All of the three major gates use a sigmoid function, \(\sigma\), to push the values into the range of 0 to 1, in order to control what gets read, written and reset (0 nothing, 1 everything). The same reason applies to the use of the hyperbolic tangent function, tanh, in the computation of the hidden state vector and the cell state candidate vector, but in this case it is so that the values range between -1 and 1. The distribution of the gradients prevents the vanishing gradient problem.

The dimensions of the vectors are as follows:

\(i_t, h_t, b_x, \tilde{C}_t, f_t, C_t, o_t \in \mathbb{R}^{h}\)

\(x_t \in \mathbb{R}^{d}\)

\(W_x \in \mathbb{R}^{h \times d}\)

\noindent where \(h\) is the number of input features (the size of the input vector \(x_t\)) and \(d\) is the number of hidden units (size of the vector \(h_t\)).

\subsection{Transformer}
The main problem with LSTM networks is that they train very slowly --- even slower than standard RNNs, given their complexity. The problem stems from the very long gradient paths in the back-propagation process --- the gradients have to be propagated all the way from the end to the start of the network and, when the input sequences are very long, this translates into a very deep network that is very slow to train. Another problem is that the input of an LSTM network is serialized, i.e., the state of an LSTM cell depends on the state of the previous one, so the previous input has to be introduced first. This implies that it is not possible to take advantage of GPUs to accelerate training with parallel computation. 

In order to deal with these problems the Transformer network architecture was created \cite{vaswani2017attention} and became the new go-to for sequence modeling.
Contrary to RNNs, Transformers ditch the recurrence completely during training and rely solely on attention. The attention mechanism helps the network take into consideration the relation between each element of the sequence, regardless of where these are positioned, eradicating the problem of long-ranged dependencies. Furthermore, Transformers suffer less from the vanishing/exploding gradient problem since it does the computation for the whole sequence simultaneously. Moreover, training a Transformer is faster as it achieves the same or better results without being as deep as an RNN in general and allows parallel computation during training.

\begin{figure}
\begin{center}
\includegraphics[width=170pt]{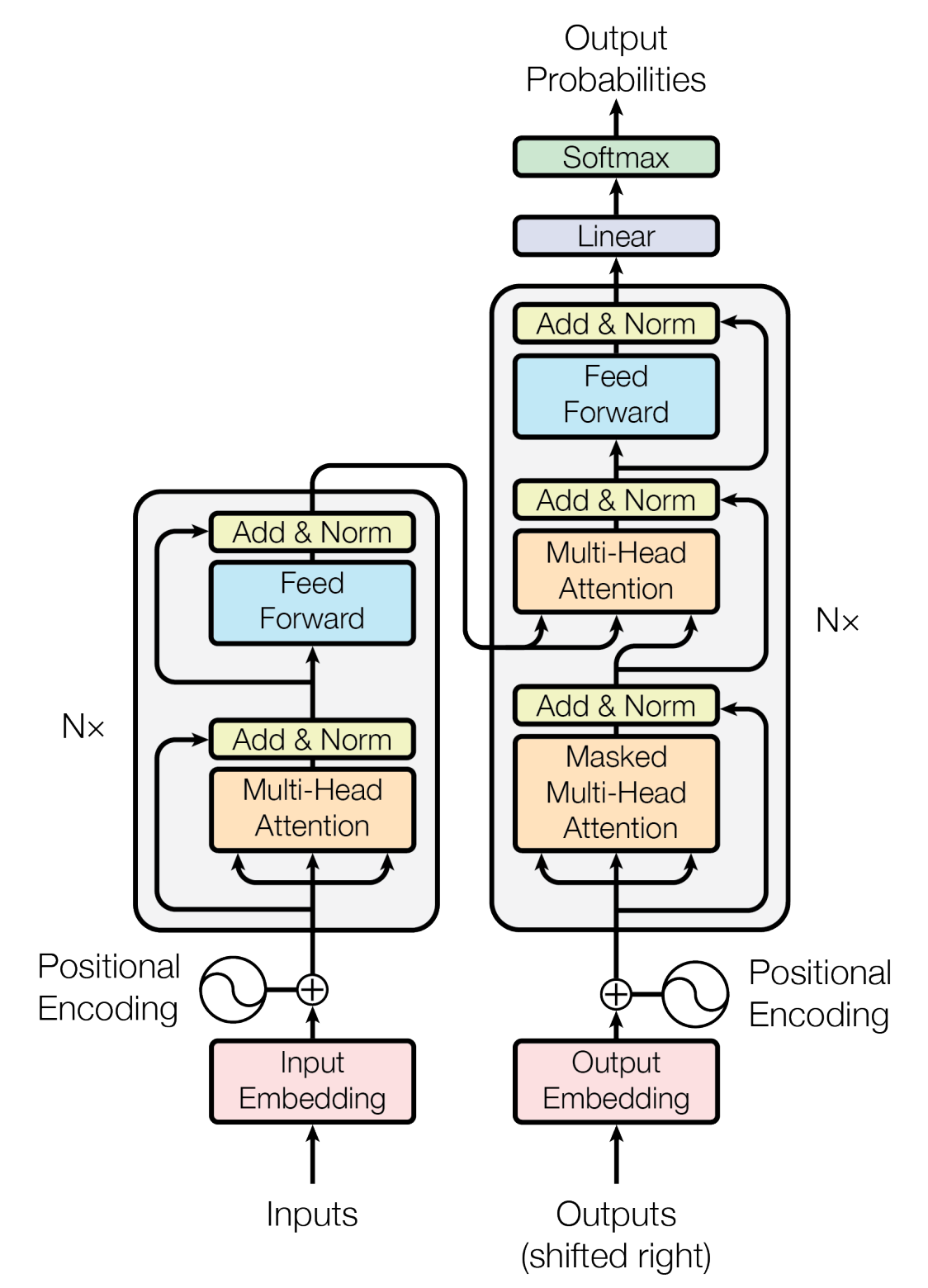}
\end{center}
\caption{The Transformer Model Architecture. \cite{vaswani2017attention}}
\label{transformer}
\end{figure}

A Transformer is a attention-based encoder-decoder model \cite{vaswani2017attention}. The encoder's objective is to encode the input into a continuous representation that holds the learned information for that entire sequence. It can be structured into several components that are displayed in Figure \ref{transformer}. As no recursion is used, the Transformer uses embeddings with positional information associated to be able to know where each sequence element is relative to each other. Moreover, the last two layers allow the Transformer to output the most probable next sequence element.

A Transformer has two Positional Embedding blocks, one feeding it's output to the encoder and another feeding it to the decoder. Each block contains two layers: Embedding layer and Positional Encoding layer. The Embedding layer transforms the input, e.g. tokens in sentence, into an embedding (vector representation). The paper of the original Transformer uses learned embeddings. In the encoder side the input for this layer are the tokens of the input sentence, while in the decoder side the input are the previous tokens generated by the decoder. RNNs know the sequence elements positions because of the recurrence, however this does not happen for the Transformer. As such the transformer model determines the order of the sequence by injecting information regarding the relative position of the sequence elements. This is done in the Positional Encoding layer by adding positional encoding to the input embedding with sine and cosine functions of different frequencies:
\begin{equation}\begin{aligned}
P E_{(p o s, 2 i)} &=\sin \left(p o s / 10000^{2 i / d_{model}}\right)
\end{aligned}\end{equation}
\begin{equation}\begin{aligned}
P E_{(p o s, 2 i+1)} &=\cos \left(p o s / 10000^{2 i / d_{model}}\right)
\end{aligned}\end{equation}
where $pos$ is the position and $i$ is the positional embedding entry. As such, each entry of the positional encoding corresponds to a sinusoid.
Using the sinusoids allows the model to extrapolate to sequence lengths longer than the ones encountered during training.

\begin{figure}[ht]
 \centering
 \includegraphics[scale=0.27]{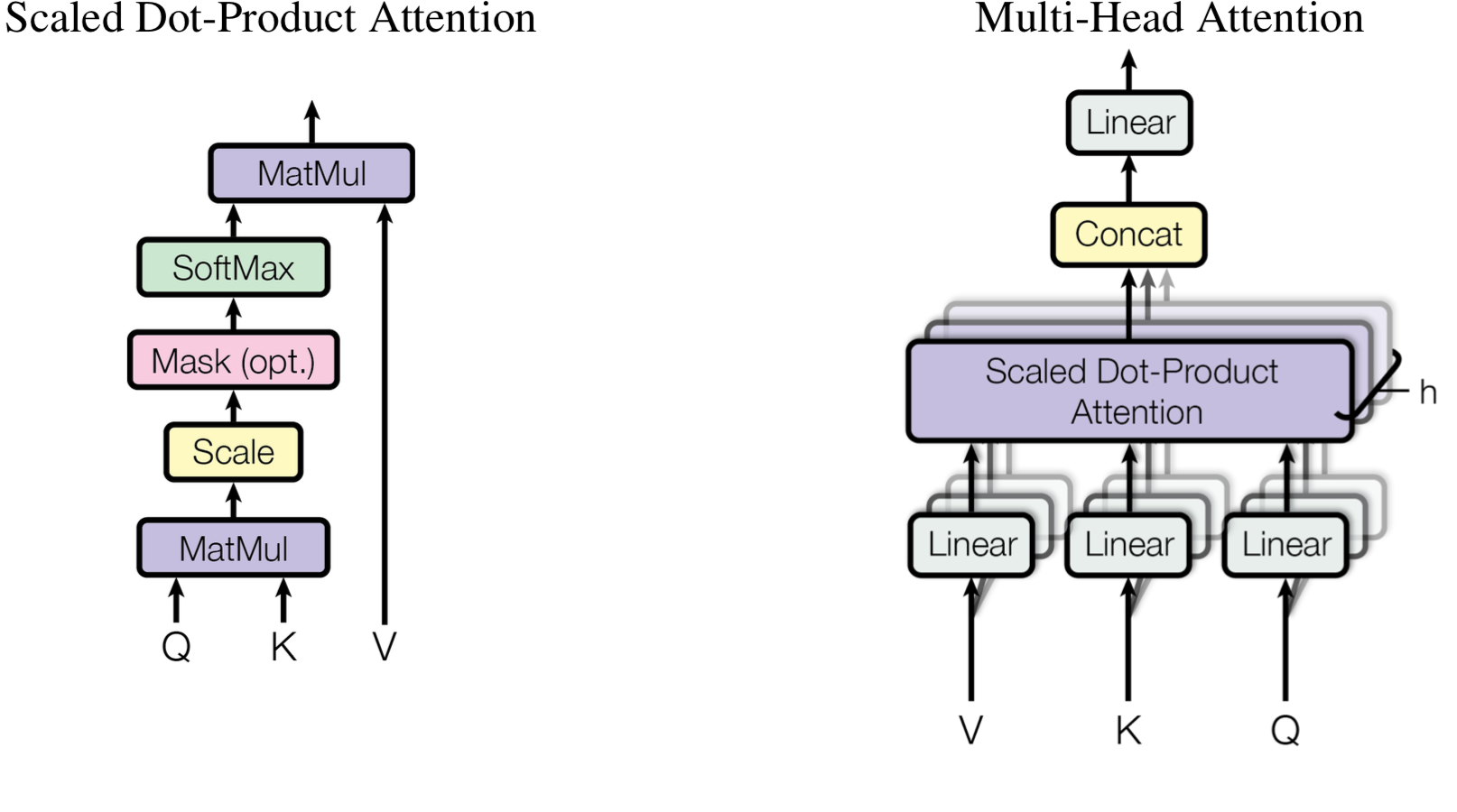}
 \caption{Scaled Dot-Product Attention (right) and Multi-Head Attention (right).  \cite{vaswani2017attention}}
 \label{fig:transformer_attention}
\end{figure}

The encoder contains two sub-layers: the Multi-Head Attention sub-layer and a Feed Foward sub-layer, as represented in Figure \ref{transformer}. To understand the Multi-Head Attention sub-layer we need to understand self-attention. The self-attention mechanisms is inspired by the retrieval of a value $v_i$ for a query $q$ based on a key $k_i$ in a database. In a database, we issue a query $q$ to a database containing a table with the keys $K$ and corresponding values $V$. The database will see which key $k_i$ is more similar to the the query $q$ and return the corresponding value $v_i$. The Transformer follows this idea, however, instead of returning a value it returns a weighted combination of the values for which the keys have an high similarity/compatibility with the queries.

\begin{equation}
\operatorname{Attention}(Q, K, V)=\operatorname{Compatibility}(Q, K) V
\end{equation}

\begin{equation}
\operatorname{Compatibility}(Q, K)=\operatorname{softmax}\left(\frac{Q K^{T}}{\sqrt{d_{k}}}\right).
\end{equation}
\noindent
Here $Q$, $K$ and $V$ are matrices, $Q$ contains the queries and $K$ contains the keys. The keys and the queries have dimension $d_k$. The matrix $V$ contains the values and each value has dimension $d_v$. This mechanism compares each token with all the other tokens and returns a matrix containing information regarding how much each word is related/compatible with the others. If we stack another encoder we will obtain matrices that tell us how much each pair of tokens is compatible with the other pairs and so on.
The transformer paper calls their self-attention mechanism as \textit{Scaled Dot-Product Attention} because of the softmax input. The mechanism is represented in Figure \ref{fig:transformer_attention} (left) alongside Multi-Head Attention (right). There is an extra Mask layer, but this layer is not used inside the encoder, it is instead used in the first decoder's sub-layer as it will be explained further ahead.

The Multi-Head Attention sub-layer is very similar to self-attention. In this case each query, key and value is linearly projected $h$ times with learned projections into different vectors. On each of these projected versions of
queries, keys and values are then performed the attention function in parallel, yielding $d_v$-dimensional output values. Each self-attention process is a head and each head returns a vector that is then concatenated into a single vector before going into the final layer. The main idea behind this is that by having multiple heads is that each head is capable of learning different things - a mechanism akin to using multiple filters in a convolutional neural network - allowing the model to have a more powerful representation model.

After this, the output of the Multi-Head Attention sub-layer is added to the original input of the encoder due to a residual connection and then the result goes through layer normalization. Then, the normalized output goes through a 
pointwise feedforward layer (the Feed Foward sub-layer) with RELUs, which is then added to the previous normalized output due to a residual connection again. The residual connections help the network during training because they allow the gradients to flow through the networks directly. The layer normalization stabilize the network which substantially reduces the duration of training. The pointwise feedforward layer is used to project the attention outputs potentially giving it a richer representation.

The decoder is structured in a similar way to the encoder. As we can see in Figure \ref{transformer}, the encoder contains three sub-layers each with a residual connection and capped off with layer normalization. The positional emebeddings are fed to the Masked Multi-Head Attention sub-layer which works similar to normal Multi-Head Attention. The difference is, since the decoder generates one token each timestep, we need to prevent the generated text to depend on future tokens, e.g. if we want to generate the sentence \textit{"I ate pie."}, the token \textit{"I"} should not depend on the tokens \textit{"ate"} or \textit{"pie"}. To do this, the authors apply a look ahead mask (refer to Figure \ref{fig:transformer_attention}) i.e. set all values in the input of the softmax which correspond to illegal connections (e.g. future tokens) to $-\infty$. The mask has to be applied to the softmax input because if we masked the softmax output the mask would put some probabilities to zero and then the probabilities would not add up to one. Afterwards, the sub-layer output is added to the original decoder's input and the result is normalized. Next, the information is passed to the second sub-layer that operates the same as the Multi-Head Attention sub-layer in the encoder. The only difference is that in this case the queries and the keys are the encoder's output and the values are the result from the last sub-layer after addition and normalization. Finally, just like in the encoder, the information is passed to a point-wise feedfoward sub-layer, followed by a residual connection and layer normalization, which marks the end of the decoder.  After the decoder is finished it's output passes through a linear layer that acts as a classifier, and then though a softmax to get the sequence element (e.g. a token) probabilities, refer to Figure \ref{transformer}.

One thing that might not be obvious is that there's really no recursion (at training time). As previously said, the decoder's input is the previously generated output, however, during training we have access to the input and the target output and as such we feed the decoder with the target output instead of using the sequence elements truly generated by the model, i.e. we assume the model generated the correct token and feed it the target output as if it was generated by the transformer. The decoder receives the whole target output and the mask masks out the invalid connections.

\subsection{BERT}
BERT, which stands for Bidirectional Encoder Representations from Transformers \cite{bert}, is a language representation model that was built upon the transformer architecture and developed to help with such tasks. BERT is different from other language representation models in the sense that it uses a pre-training technique that relies on bidirectional modeling, as opposed to a unidirectional modeling (left to right or right to left). Until BERT appeared, language representation models would usually train this way since the existence of a direction was what allowed the system to generate a probability distribution of the words. 

When we're given the task of understanding textual data, such as a sentence, we subconsciously understand each word by looking at its surrounding words. That is the ultimate way that we have to know in which context it is being applied. This type of understanding is bidirectional. And so, given that language understanding is a bidirectional task for humans, it only makes sense that it should too for computers. However, it is not as straight-forward to put in motion.

\begin{figure}
\includegraphics[width=\textwidth]{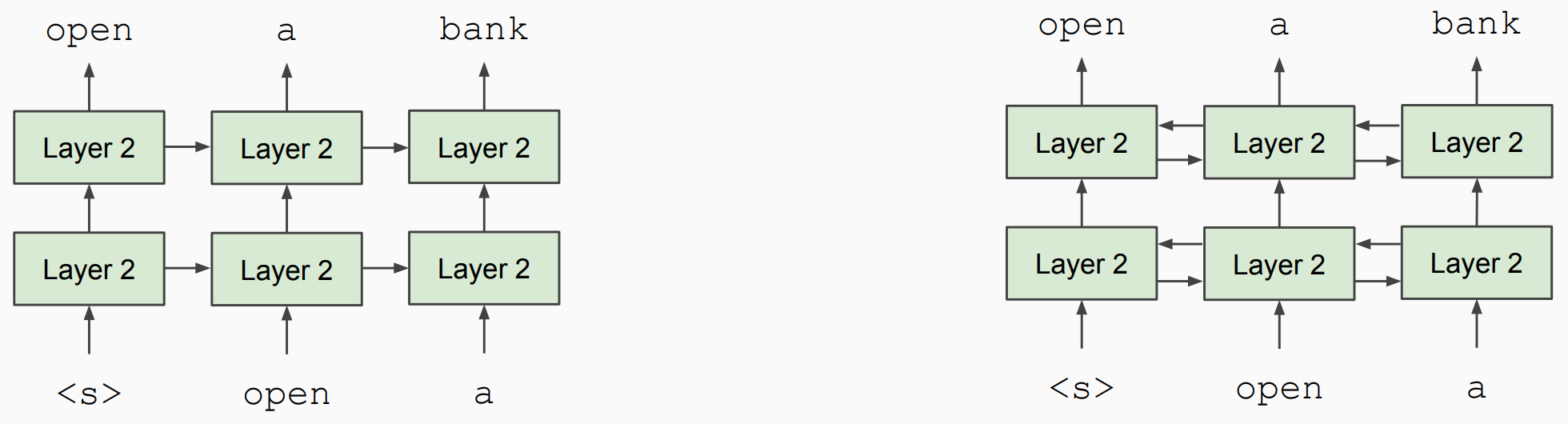}
\caption{Unidirectional and Bidirectional training, on the left and right, respectively. \cite{bert}}
\label{bert}
\end{figure}

The problem that emerges when applying bidirectional unsupervised training is the trivial prediction of words. In a unidirectional training, words can only see their left or right context. It is the existence of direction in the training that allows for an unbiased prediction of each next word. When that does not exist, as is the case in a bidirectional training, the prediction of each word --- based on both its left and right words --- becomes biased, and words are then able to predict themselves. (See Figure \ref{bert}).

BERT manages to treat this issue by using a masking technique. In a normal pre-training cycle, BERT (which relies on unsupervised training) goes through all of the unlabeled text from which it is learning from and, for 15\% of the tokens in the text, it applies a mask (by replacing those tokens with a [MASK] token). Only these masked tokens are to be predicted.

By masking a small percentage of tokens, the inevitability of predicting a word that was there to see all along during training disappears. Since the prediction is only done on masked tokens, and not on the entire text input, it creates a mismatch between the pre-training and the fine-tuning, where there is no masking due to it being supervised training. To solve this issue, the masking strategy is not only focused on replacing tokens with a [MASK] token --- this is done on 80\% of the masked tokens (80\% of 15\% = 12\% of the entire text). For another 10\% of the tokens (1.5\% of the entire text) they get replaced with another random token in the text, and in the remaining 10\% (1.5\%) it just leaves the original token unchanged. The purpose of this last percentage is to create a bias towards the actual observed word so that the system can actually learn the correct prediction.

The masking is applied to 15\% of the tokens since this was the number that achieved a better performance. The problem with a small percentage of masked tokens is that the learning curve gets much bigger. On the other hand, a greater percentage of masked tokens results in a lack of context, making it harder to predict a word by its surroundings.

Another task for which BERT also pre-trains is the prediction of next sentences. This is especially important to determine the relation between two sentences. It becomes relevant in tasks such as Question Answering and Natural Language Inferencing.

In order to represent the words that the system receives as input, BERT creates word embeddings based on three components. The first component is the token embedding. This is the value that is assigned to the token, in a fixed vocabulary that BERT uses. The second, Segment Embedding, is simply an embedding of the segment to which the word belongs. These are not to be confused with the default sentence embeddings that BERT creates in the embedding for the [CLS] token that appears at the start of every sentence and collects information about the word tokens in the sentence by means of average pooling. Finally, we have the position embedding, which is precisely what its name states, an embedding of the word position in the input.

\begin{figure}
\includegraphics[width=\textwidth]{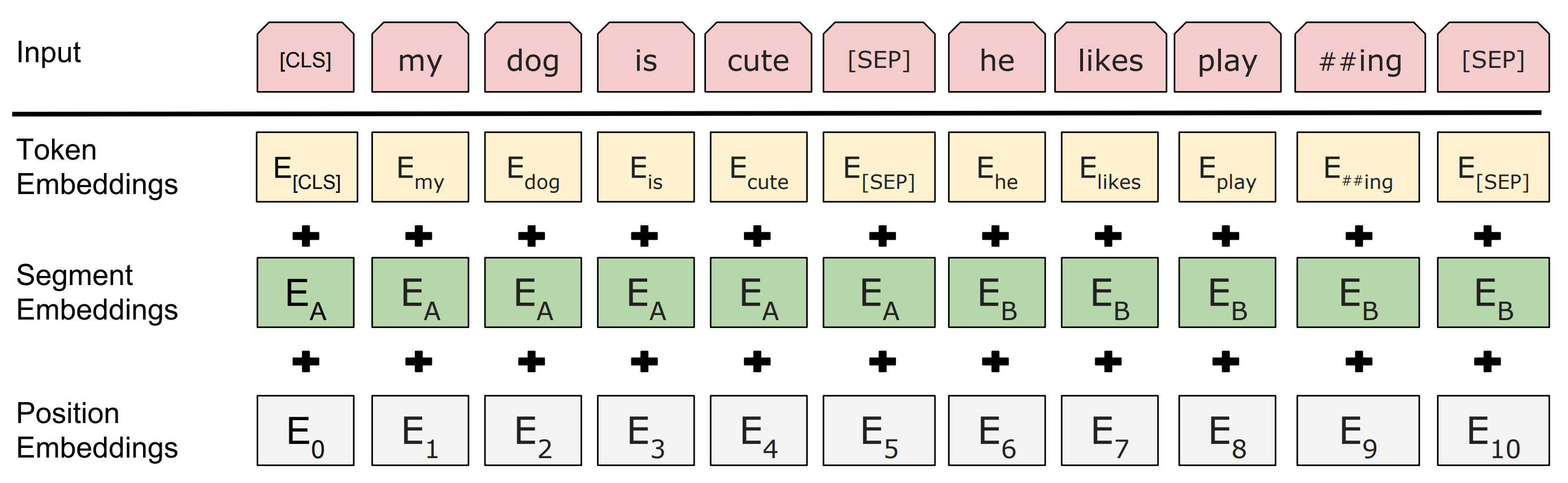}
\caption{The three components of word embeddings - Token, Segment and Position. \cite{bert}}
\label{bert2}
\end{figure}

And so, the input embeddings are the sum of the token embeddings, the segment embeddings and the position embeddings. Figure \ref{bert2} describes an example of the embedding of a sentence.

These embeddings are then fed to the neural network by the encoder, and its outcome is then received by the decoder.

\subsection{GPT2}
The GPT2 (Generative Pretrained Transformer 2) model \cite{radford2019language} is a large transformer based language model and is able to be used for any problem that can be modeled as a language modeling problem (e.g. machine translation, text summarization, question-answering). The model can create coherent paragraphs of text and achieve state-of-the-art performance on many language modeling benchmarks. Moreover, GPT2 performs rudimentary reading comprehension, machine translation, question answering, and summarization without task-specific training, i.e. without having to be specifically trained to do the task.
To achieve these results, GPT2 was trained in a unsupervised manner on a large 40GB dataset called containing over 8 million documents.

\begin{figure}[ht]
 \centering
 \includegraphics[scale=0.20]{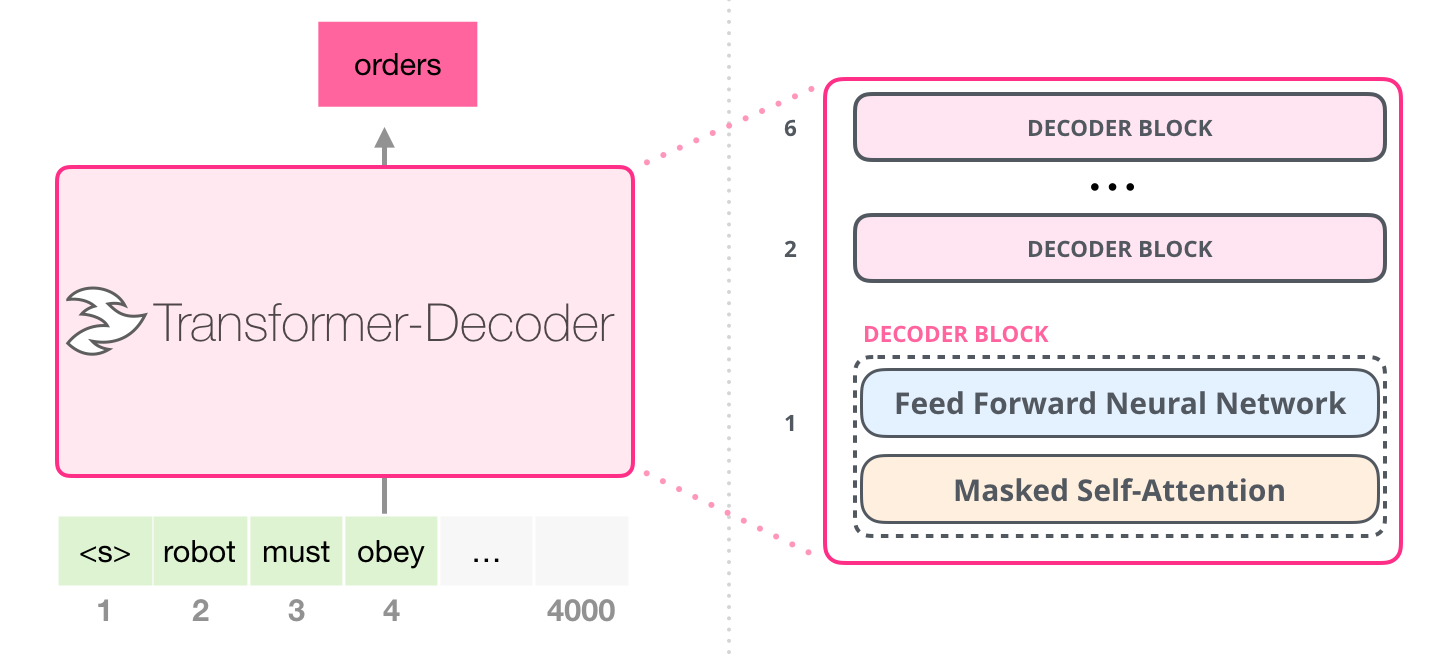}
 \caption{Structure of a transformer using decoder blocks only. These blocks are used in GPT2. Model dimention is the dimention of the embeddings. The <s> token is a start token inserted at the begining of every input. Layer normalization is not represented. From jalammar\protect\footnotemark.}
 \label{fig:transformer-decoder}
\end{figure}

There is an important different between GPT2 and the original transformer architecture. While vanilla transformers have both encoder and decoder blocks, GPT2 relies solely on decoder blocks stacked together. Moreover, the decoder blocks used in GPT2 do not use the second self-attention layer (the one connected to the original transformer encoder block) as represented in Figure \ref{fig:transformer-decoder}.

\footnotetext{http://jalammar.github.io/illustrated-gpt2/\#part-1-got-and-language-modeling}

Just like the previous neural models, tokens need to be encoded before being fed to the model. The model uses a vocabulary of 50,257 byte-pair-encoding (BPE) tokens. BPE allows the model to encode any word, given that is written in a Unicode string, so GPT2 can be evaluated on any dataset regardless of pre-processing, tokenization, or vocabulary size.

GPT2 uses absolute positional embedding instead of the relative ones used in the original transformer architecture. This means that the absolute position of the token is encoded in the embedding.

Another important aspect to GPT2 is how decoding is done. 
In the paper \textit{top-k sampling} is used with $k = 2$ which reduces repetition and encourages more abstract summaries than greedy decoding. This means that at the end of the decoder, instead of choosing the most probable token with greedy sampling, the token probability distribution is truncated to the top $k$ tokens and then re-normalized.

\subsection{GPT3}

Recently, Brown et al (\cite{brown2020}) introduced an even larger language model, with 175 billion parameters, and tested its performance in the few-shot setting. They pre-trained the model in a similar way to the GPT2 model, but then did not fine-tune it for specific tasks. Instead, GPT3 was evaluated under 3 different conditions:
\begin{itemize}
    \item ``few-shot learning", or in-context learning where between 10 and 100 demonstrations are given to the system;
    \item ``one-shot learning", where only one demonstration was given to the system;
    \item ``zero-shot” learning, where no demonstrations were given to the system and only
an instruction in natural language was allowed;
\end{itemize}

GPT3 achieved near state-of-the-art results in many NLP tasks, even competing with fine-tuned models. Its Architecture is very similar to GPT2's architecture, just differing in the attention layers. It was trained on 300 bilion tokens and the training computation power spent was over 2000 Pentaflop/s-days.

Despite its successes GPT3 also showed some limitations. One is common to only self-supervised trained language models, which is that task specification has to be done as a prediction problem, when a legal cognitive assistant might be better thought of as taking goal-directed actions to help the user, rather than just making predictions.

\section{Semantic Representation Languages}
This section presents an overview in Semantic Representation Languages and Semantic Information Extraction in Natural Language Processing, to help the reader understand how the use of additional semantic information extracted from text can then be later used to enhance systems for tasks such as information retrieval or norm extraction.

\subsection{Semantic Roles}
The simplest form of semantic information that can be extracted from sentences is named semantic roles or thematic roles. Thematic Roles \cite{srl2019} are abstract models of the role an argument plays in the event described by a predicate. These roles allow natural language understanding systems to better understand how the participants relate to events, i.e. it helps answer the question “Who did what to whom” and possibly “where” and “when”. Some popular thematic roles are shown in Table \ref{table:thematic_roles}.

\begin{table}[]
\footnotesize
\centering
\begin{tabular}{|l l|}
\hline
\textbf{Thematic Role} & \textbf{Definition}                                 \\
\hline
Agent                  & The volitional causer of an event                   \\
Experiencer            & The experiencer of an event                         \\
Force                  & The non-volitional causer of the event              \\
Theme                  & The participant most directly affected by an event  \\
Result                 & The end product of an event                         \\
Content                & The proposition or content of a propositional event \\
Instrument             & An instrument used in an event                      \\
Beneficiary            & The beneficiary of an event                         \\
Source                 & The origin of the object of a transfer event        \\
Goal                   & The destination of an object of a transfer event   \\
\hline
\end{tabular}
\caption{Common thematic roles. From \cite{srl2019}.}
\label{table:thematic_roles}
\end{table}

There are multiple models of Thematic Roles that may use different sets of roles. Some sets have a great number of roles while others may contain a smaller more abstract set. Most thematic role sets have about a dozen roles, there are also smaller sets with even more abstract meanings. On the other hand, there are also sets with a lot of roles that are specific to situations, like Proposition Bank (refer to Section \ref{sec:propbank}). Any set of roles, regardless of size of role abstraction is called a semantic role.
Computational systems use semantic roles as shallow meaning representation to make simple inferences that are not possible to make using only the string of words or the parse tree. As such, semantic roles help generalize over different predicates. However, it is difficult to come up with a universal/standard set of roles and to also produce a formal definition for each role. For example, the words correspondent to each Thematic Roles may be in different positions depending on the sentence and do not correspond to the same dependency relations every case, e.g. the agent is not always the subject as one might expect.

\subsection{Generalized Semantic Roles}
The problems that came with using Thematic Roles led to alternative semantic role models such as the Generalized Semantic Roles. Generalized Semantic Roles \cite{dowty1991thematic} are defined using a set of heuristics that define more agent like or more patient like meanings. As such, Generalized Semantic Roles define PROTO-AGENT to represent agents and PROTO-PATIENT to define who/what suffers the action.

Arguments that follow the PROTO-AGENT role have the following properties.
\begin{itemize}
    \item Volitional involvement in the event or state
    \item Sentience (and/or perception)
    \item Causing an event or change of state in another participant
    \item Movement (relative to the position of another participant)
    \item Exists independently of the event named by the verb
\end{itemize}
Arguments that follow the PROTO-PATIENT role have the following properties.
\begin{itemize}
    \item Undergoes change of state
    \item Incremental theme
    \item Causally affected by another participant
    \item Stationary relative to movement of another participant
    \item Does not exist independently of the event, or not at all
\end{itemize}

These Generalized Semantic Roles are used by The Proposition Bank \cite{palmer2005proposition} and by FrameNet \cite{ruppenhofer2006framenet}.

\subsection{Proposition Bank} \label{sec:propbank}
The Proposition Bank (PropBank) \cite{palmer2005proposition} is a resource of sentences annotated with semantic roles. The English PropBank has all sentences from Penn TreeBank labeled, however there are more languages available (e.g. Portuguese).
Each verb contains a specific set of roles, as it is very difficult to generalize roles that fitted every verb, that are divided into two groups: the numbered arguments and the modifier arguments.
These arguments are used to structure each predicate. They are outlined in a specific frame file for that predicate.
Although the arguments are specific to each verb, generally they correspond to certain semantic roles \cite{propbank_guidelines}.
The common semantics of each PropBank argument are shown in Table \ref{table:propbank_arguments}.

\begin{table}[]
\footnotesize
\centering
\begin{tabular}{|ll|}
\hline
\textbf{Arguments} & \textbf{Semantic Roles} \\
\hline
ARG0      & agent                                 \\
ARG1      & patient                               \\
ARG2      & instrument, benefactive, attribute     \\
ARG3      & starting point, benefactive, attribute \\
ARG4      & ending point                           \\
ARGM      & modifier\\                         
\hline
\end{tabular}
\caption{List of arguments in PropBank. From \cite{propbank_guidelines}.}
\label{table:propbank_arguments}
\end{table}

There are a set of numbered arguments associated with each verb. The most important are ARG0 and ARG1. Typically, ARG0 are  the  subjects of transitive verbs and a class of intransitive verbs. This argument has PROTO-AGENT properties as defined in the previous section. The argument ARG1 are the objects of the transitive verbs and the subects of intrasitive verbs. These arguments have PROTO-PATIENT qualites.

Another important set of arguments are the ARGM. These represent modification or adjunct meanings.  These are relatively stable across predicates, so there is no need to list them for every verb frame. PropBank has a variety of ARGM arguments. Some of the available modifiers are the following.

\begin{table}[]
\footnotesize
\centering
\begin{tabular}{|l l|}
\hline
\textbf{ARGM}     & \textbf{Semantic Roles} \\ 
\hline
TMP      & indicates when the event happened.                           \\
LOC      & where the event took place.                                   \\
ARGM-COM & indicates with who an action was done (besides the ARG0).      \\
ARGM-NEG & used for markes of negation (e.g. "no", "n't", "never", "no longer") \\
ARG4     & ending point                           \\
ARGM     & modifier \\
\hline
\end{tabular}
\caption{List of ARGM arguments in PropBank. From \cite{propbank_guidelines}.}
\label{table:propbank_ARGM_arguments}
\end{table} 

\subsection{AMR}
The abstract meaning representation language (AMR) \cite{banarescu2013} consists in a rooted, directed, acyclic graph with labeled edges (relations) and nodes (concepts) expressing "who is doing what to whom". It is relatively easy to read by human readers and easily processed by computational processes. Translating sentences in natural language to an AMR representation is simpler than doing it for more formal representation languages such as First Order Logic. However, Johan Bos \cite{Bos2016} has shown that AMR can achieve the expressive power of First Order Logic with a couple of simple extentions to the language.

AMR has received some attention from the research community. For instance, Sheng Zhang and colleagues worked on a automated parser for AMR \cite{zhang-amr2019}, where they treat AMR parsing as a Sequence-to-Graph transduction problem. Their system is composed by two main modules, a pointer-generator network is used for node prediction (detecting the AMR concepts in the sentence), while a deep biafine classifier uses the hidden states of the decoder as input to predict  edges (relations between nodes). 

Another example of a deep AMR parser is the ConvAMR system, developed by Viet et al. \cite{Viet2017}. Instead of a encoder-decoder based model, they propose a convolutional sequence-to-sequence model with a graph linearization technique. Of important notice is the fact that ConvAMR was actually trained to parse an AMR representation from legal text, more specifically an English version of the Japanese Civil Code. Table \ref{table:legal_AMR} shows an example of a sentence and its corresponding AMR representation in PENMAN notation.  

\begin{table}[]
\footnotesize
\centering
\begin{tabular}{|p{10.0cm}|}
\hline
\textit{Unless otherwise provided by applicable laws, regulations or treaties, foreign nationals shall enjoy private rights.}\\
\hline
\begin{verbatim}
(e / enjoy-01
    :ARG0 (n / national
        :mod (f / foreign))
    :ARG1 (r / right-05
        :ARG1-of (p / private-02))
    :condition (p2 / provide-01 :polarity -
        :OR (o / or
            :op1 (l / law
                :mod (a / applicable))
            :op2 (r2 / regulate-01)
            :op3 (t / treaty))))
\end{verbatim}\\
\hline                              
\end{tabular}
\caption{Example of legal sentence and its AMR representation. From \cite{Viet2017}}
\label{table:legal_AMR}
\end{table} 

\subsection{An approach to Open Information Extraction using deep learning}

\begin{figure}
\includegraphics[width=\textwidth]{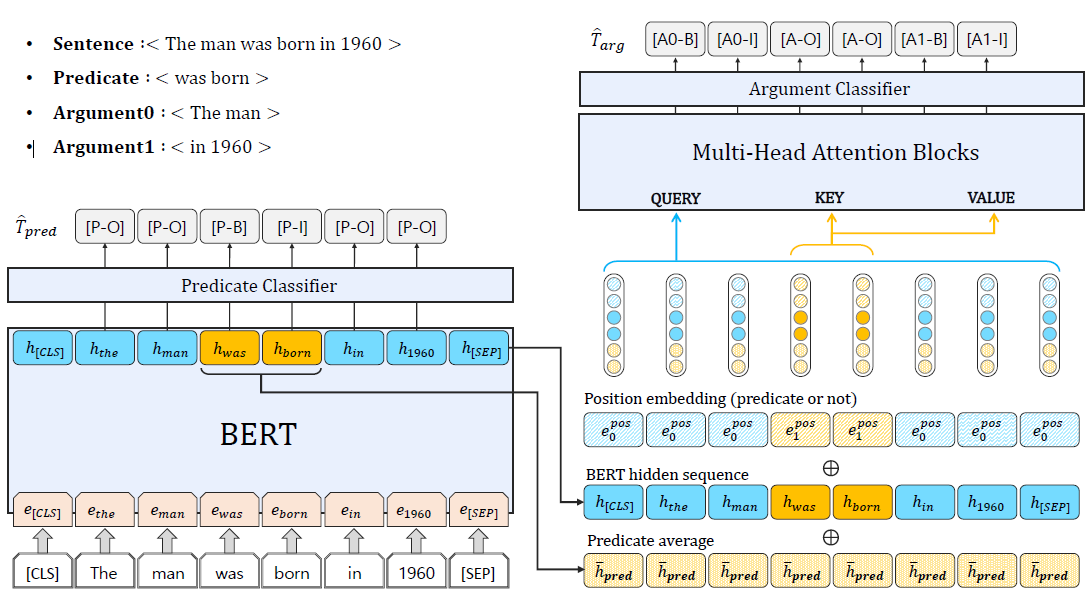}
\caption{Architecture of Multi$^2$OIE. Predicates are extracted using the hidden states of BERT, and then the hidden sequence, average vector of predicates, and position embedding are concatenated and used as inputs for multi-head attention blocks for argument extraction. From \cite{ro2020})}
\label{fig:Multi2}
\end{figure}

Open Information Extraction (OIE) is a task very similar to shallow semantic representation. It attempts to extract a sentence's predicate and the main arguments for it. Recently, Ro et al (\cite{ro2020}) presented a Multi-language BERT-based architecture for this task (see Figure \ref{fig:Multi2}), which they named Multi$^2$OIE.
Their model surpassed previous state-of-the-art architectures in the Portuguese language. In their proposed architecture sentences are tokenized
by SentencePiece (\cite{kudo2018}) and BERT is used to label the tokens that constitute the predicate, with its result then fed to a series of Multi-Head attention blocks that extract the Arguments.

For Portuguese, Multi$^2$OIE obtained an F1 score of 59.1, with Precision 56.1 and Recall 62.5 on the binary extraction, which was well below that obtained for English, but never-the-less better that for instance Predpatt applied to the same tasks. These results were obtained without specific training for the Portuguese language, which means that they can possibly be improved with specific training.

\section{Applications of NLP to the Legal Domain}
Researchers have started exploring the recent advances in NLP achieved with deep learning models to a set of downstream tasks. In this section we will address examples of systems that illustrate how Natural Language Processing can be used to improve a set of tasks that will be useful for the legal domain, such as information retrieval or norm extraction. We will depict examples of systems that were directly applied to legal texts, but also of systems whose techniques could easily be applied and transposed to a legal domain.

\subsection{Semantic NLP Approaches to Legal Information Retrieval}
The field of NLP has revealed to be quite relevant in the creation of new ways of interpreting and representing legal text so that normal citizens with no introduction to fundamental law concepts can understand and query the system. For this complex task, there are a few factors that weigh in the virtuous operation of a NL-based law query system.

The order of the words in a sentence is a very important factor in its semantic value. The first few sentence encoders that were released - some that are essentially repurposed word encoders - have fallen into the same predicament. The lack of word order encoding. Two sentences with the exact same words, but different ordering, can have totally different meanings. And the context defined by that order can influence the meaning of a specific word. For instance, when an individual says, \textit{I went for a run}, we assume that the word \textit{run} has a meaning related to the sport. When another individual says, \textit{You have to run this on your computer}, the meaning is completely different from the other sentence. Not only is it being used as a verb, but also the meaning of it is no longer related to the sport, but rather to a computational task. And, even in a case where the word \textit{run} was used as a verb connected to the sport, it could still be differentiated from the computational meaning due to the associated words, for most of the cases.

It is important to evaluate the Semantic Textual Similarity (STS) between two pieces of text, which is the concept of similarity of meaning between them. This task is useful for identifying how search systems are able to capture the meaning of a query and use it as a factor in the selection of the search results. One important benchmark to measure the performance of models in the task of STS is the GLUE benchmark. It uses metrics and STS datasets specifically tailored for the evaluation of a model's capacity to encode meaning into the token embeddings it produces.

\subsubsection{USE}
Google’s USE (Universal Sentence Encoder) \cite{use} was one of the language models that were created to mitigate the issue stated above. able to consider the order between words. In USE’s case, there are two models available. The first, Deep Averaging Network (DAN), averages together the input embeddings for words and bi-grams and then passes them through a feed-forward Deep Neural Network (DNN) to produce the sentence embeddings. By considering bi-grams, it allows the representation of pairs of words in a sentence embedding. 

The second one, Transformer, constructs sentence embeddings using the encoding sub-graph of the transformer architecture. It uses attention to compute context aware representations of words in a sentence. These representations are then converted to a fixed length sentence encoding vector. 

The Transformer model is much more complex than the DAN one and more resource demanding. It does have a slightly higher accuracy but the DAN model achieves a computation time that is linear to the length of the input sequence. USE is also available in a version pre-trained for Portuguese (Multilingual) \cite{muse} in the Transformer variant and in a Convolutional Neural Network (CNN) version with a reduced accuracy.

\subsubsection{InferSent}
Another very popular sentence encoder model is InferSent \cite{conneau2017supervised}. This model has proven to be better at certain key aspects of sentence representations, when compared to USE --- more specifically in terms of distinguishing between a sentence and its negation. USE, however, performs better when estimating the STS between embeddings and also seems to excel at detecting relevant word order differences.

\subsubsection{Sentence-BERT}
\begin{figure}
\includegraphics[width=\textwidth]{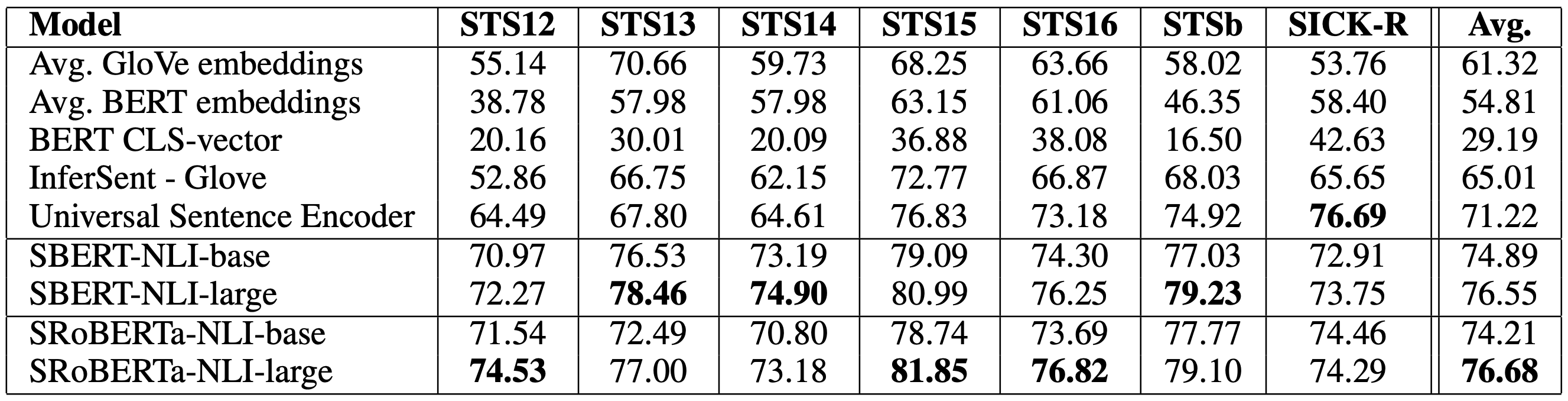}
\caption{Spearman rank correlation  \textit{c} between the cosine similarity of sentence representations and the gold labels (label that classifies two sentences on the basis of their similarity) for various STS tasks. Performance is reported by convention as \textit{c} x 100. STS12-STS16: SemEval 2012-2016, STSb: STSbenchmark, SICK-R: SICK relatedness dataset. \cite{reimers2019sentencebert}}
\label{s-bert}
\end{figure}

A team of researchers from UKP (Ubiquitous Knowledge Processing Lab) developed a system, Sentence-BERT \cite{reimers2019sentencebert}, heavily based on the BERT model that has averaged promising results in all of the SemEval editions from 2012 to 2016, especially when compared to BERT's averaged embeddings or CLS embeddings --- and even when compared to USE and Infersent. (See Figure \ref{s-bert}).

\begin{figure}
\begin{center}
\includegraphics[width=100pt]{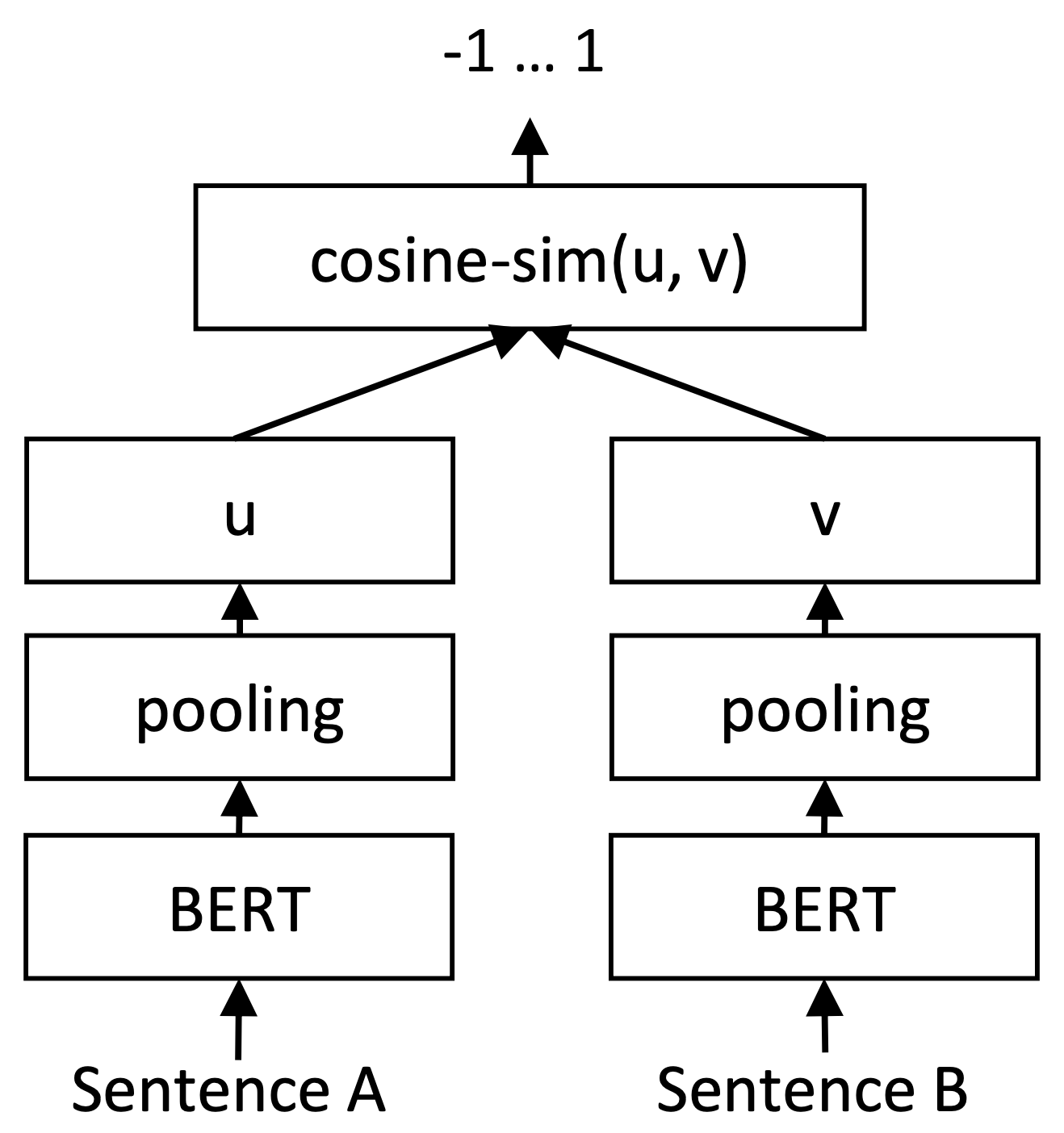}
\end{center}
\caption{SBERT architecture for STS tasks. \cite{reimers2019sentencebert}}
\label{s-bert2}
\end{figure}

Sentence-BERT, or SBERT, is essentially a modified pre-trained BERT model that uses siamese and triplet network structures in order to extract sentence embeddings that are semantically relevant. The system trains by encoding two sentences in the siamese way, i.e., running two identical networks adjusted with the same parameters on two different inputs --- in this case two BERT networks receive each a sentence to encode. These encodings are then passed through a pooling process that the team behind SBERT found to have a better performance with a mean agreggation strategy, as opposed to a max or [CLS] vector strategy. Like we have seen previously, it consists of averaging the token embeddings of a sentence. That way, all of the token embeddings that BERT outputs are joined into one vector, consequently joining a layer of the size of the sequence of tokens into one output. Finally, both sentence encodings are then compared using cosine-similarity. The whole process is depicted in the diagram in Figure \ref{s-bert2}.

The training is done using a combination of two datasets --- the Stanford Natural Language Inference \cite{bowman2015large} (SNLI) and the Multi-Genre NLI \cite{williams2017broadcoverage}. The SNLI has a body of 570,000 sentence pairs annotated with the labels contradiction, entailment and neutral. MultiNLI contains 430,000 sentence pairs and covers a range of genres of spoken and written text. 

When developing SBERT, the UKP team compared the performance of different metrics in the evaluation of each sentence pair similarity. They experimented with the most used one, cosine-similarity, but also tried to use negative Manhattan and Euclidean distances. After the experiments, they concluded that the metric used, between the three, wasn't relevant given that the results were roughly the same. Therefore, they continued to use cosine-similarity as the metric of STS. They also considered using a regression function that would map sentence embeddings to similarity scores but refrained from doing so given the resource exhaustion that would occur.

When comparing the performance of SBERT, in STS tasks, two different strategies of training were used --- Unsupervised and Supervised Learning. For the unsupervised approach, SBERT only retained the knowledge that it had gained with the pre-training from BERT (based off of Wikipedia) and NLI data. To evaluate this system three datasets were used --- STS tasks 2012-2016\footnote{http://alt.qcri.org/semeval2020/}, the STS benchmark \cite{semeval2017} and the SICK-Relatedness\cite{sick} datasets. These three datasets include labels for sentence pairs that define, on a scale of 0 to 5, how semantically related they are. SBERT was able to outperform both InferSent and USE on most of the datasets, with the exception of the SICK-R dataset, in which USE gained an edge due to its training on a variety of diverse datasets that seemed to better fit the data in SICK-R. The results can be seen in Figure \ref{s-bert}.

For the supervised learning, SBERT was fine-tuned on the training set of the STS benchmark dataset using cosine similarity as the metric for sentence embedding similarity alongside a mean squared error loss function to assess the quality of each prediction. This dataset has proven to be very popular in the evaluation of supervised datasets given the quality of the sentence pairs and its dimensions --- it is composed of 8628 sentence pairs that are divided into three categories (\textit{captions}, \textit{news} and \textit{forums}).

Apart from the fine-tuning done only on STSb, another experiment was done by training on the NLI dataset first and then the STSb. This latter option resulted in a considerable improvement. In the same paper~\cite{reimers2019sentencebert} it was also found that using RoBERTa, a BERT based language model, (instead of BERT) did not make much difference in the final results. These findings are displayed in table~\ref{s-bert3}.

\begin{table}
\begin{center}
\includegraphics[width=130pt]{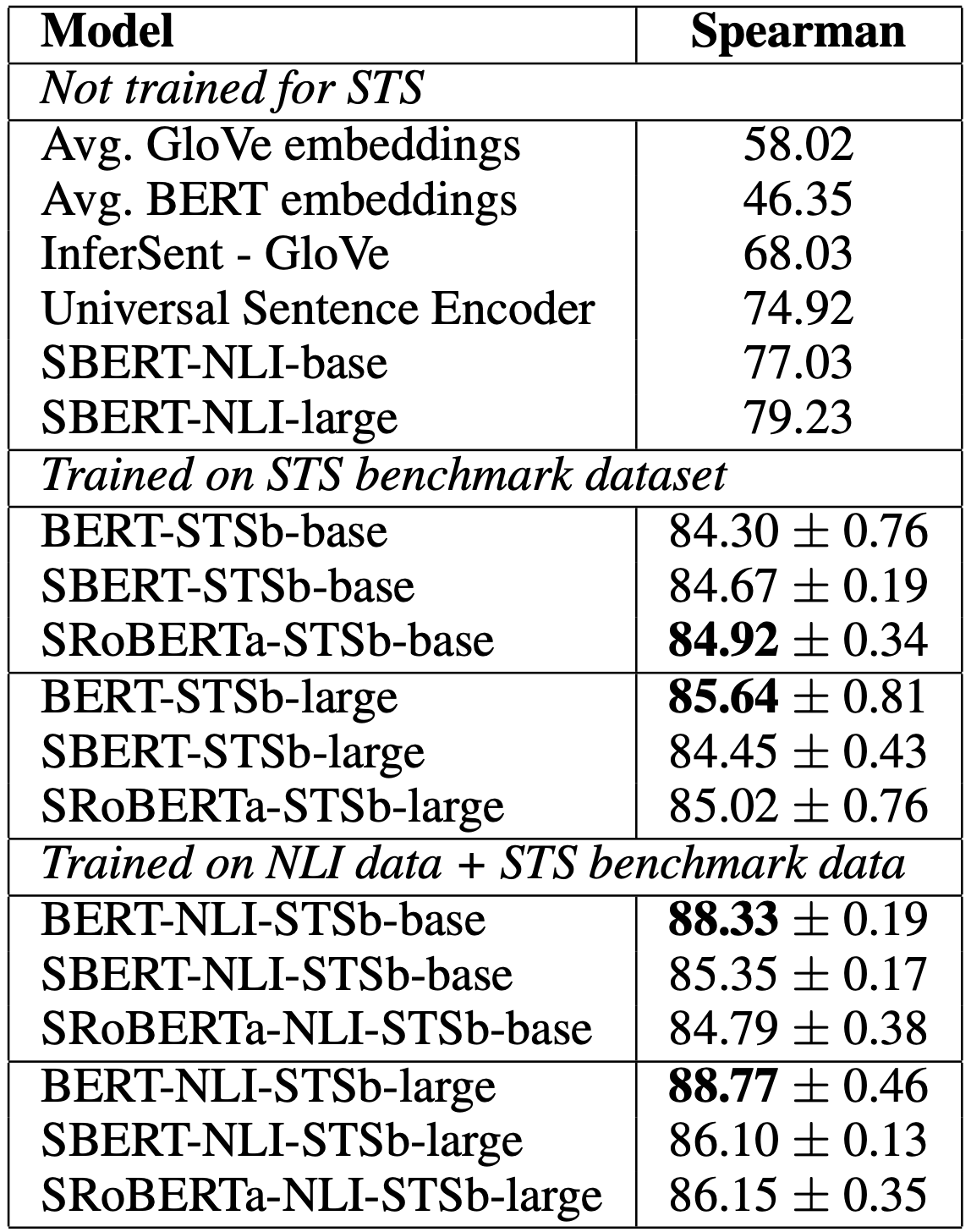}
\end{center}
\caption{Evaluation on the STS benchmark test set. SBERT was fine-tuned on the STSb dataset, SBERT-NLI was pretrained on the NLI datasets, then fine-tuned on the STSb dataset. \cite{reimers2019sentencebert}}
\label{s-bert3}
\end{table}

\subsubsection{ORQA}
Kenton Lee and colleagues \cite{lee2019} from Google Research came up with an interesting idea for creating a model for Question Answering (QA) with very limited annotated information. Although Question Answering might seem different from information retrieval (while information retrieval is more focused on syntactic and semantic matching, QA requires deeper language understanding), these tasks can benefit from the same type of information. A QA system is typically composed by two subsystems: a retriever and a reader component. The retriever component is similar to a component developed for information retrieval, given a particular question or query it needs to find the most relevant set of evidence candidates to answer it. The reader component will then determine the exact words from the evidence candidates that contain the answer.     

In their approach, named Open Retrieval Question Answering system (ORQA), Lee et al. are able to perform end-to-end learning (i.e. simulateneously train the retriever and reader on the full corpus instead of using a subselection returned by a black-box information retrieval system) by pre-training the retriever using a Inverse Cloze Task. In a Inverse Cloze Task, a sentence is treated as a pseudo-question, and its surrounding context is treated as pseudo-evidence. The assumption here is that more often than not, the surrounding context of a sentence will be semantically correlated with it, making it a good heuristic for the unsupervised creation of a dataset for training a retrieval component. 

\begin{figure}
\begin{center}
\includegraphics[width=250pt]{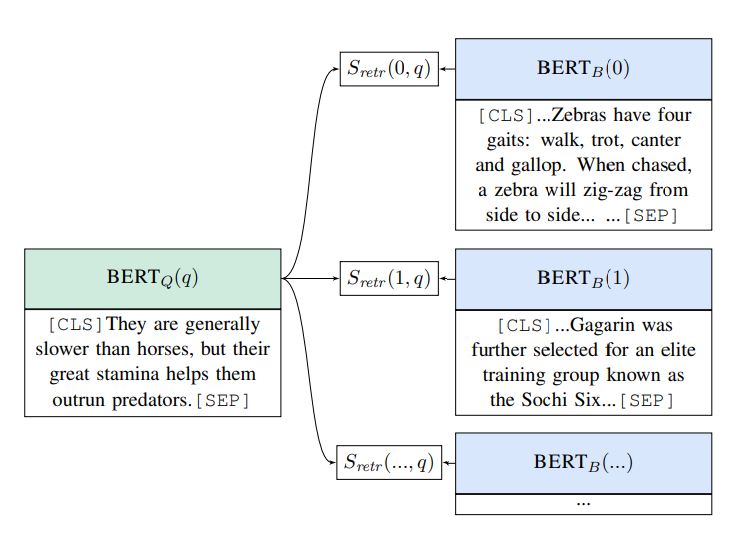}
\caption{Example of the Inverse Cloze Task. A random sentence and its context are derived from the text "...Zebras have four gaits: walk, trot, canter and gallop. They are generally slower than horses but their great stamina helps them outrun predators. When chased, a zebra will zig-zag from side to side..." \cite{lee2019}}
\label{fig:ict}
\end{center}
\end{figure}

Figure \ref{fig:ict} shows an example of the Inverse Cloze Task. Each segment from the corpus is further divided into sentences. One random sentence is selected as the pseudo-query, and the remaining text from the segment is selected as the pseudo-evidence. The retriever model is trained to select the correct context from a set of candidates in a batch. Negative candidates are selected randomly from other segments of the corpus. 

ORQA was evaluated in several different QA datasets with interesting results. In datasets where the writer of the question already knows the answer (or has some evidence for it) a traditional BM25 search performs better, as the wording of the question will contain lexical cues for the answer. However on datasets where a user is genuinely seeking information he does not know (natural question datasets), the ORQA system is able to significantly outperform BM25. We believe that this type of approach is very useful for developing an information retrieval system for the Portuguese Legal Domain, as the type of queries performed by users will be of the latter case. There is in fact a recent trend in the research community of exploring Question Generation systems as a way of quickly building large datasets of question and answers, which can then be used to train models for Question Answering, Information Retrieval, and even Chatbots. For instance, Dhole and Manning \cite{dhole2020} use syntatic and semantic rules based on SRL structure automatically parsed from text to generate questions and answers

\subsubsection{Quin and QR-BERT}
Although not applied to the legal domain, we believe that the models (QR-BERT and the reranking model) and the methodology used for training the Quin system could easily be transposed to a legal corpus. Quin \cite{samarinas2020} was recently developed with the goal of providing the most relevant passages to a particular query about Covid-19. One of the interesting aspects of Quin is that it combines a syntactic-based search mechanism, BM25, together with a latent semantic retrieval model: QR-BERT.

QR-BERT is a BERT-based model that is pre-trained on Natural Language Inference datasets, namely SNLI and MultiNLI (both referred as NLI), and FactualNLI — a dataset derived from existing question answering datasets. QR-BERT is pre-trained with the following classsification objective:

\[o = softmax(W[u; v; |u-v|])\]

\noindent where \(u\) is the embedding of the premise sentence and \(v\) is the embedding of the hypothesis sentence, \(W_{3 \times 3k}\) a linear transformation matrix, where \(k = 768\) is the dimensionality of the hidden representation of BERT-base, and \([u; v; |u - v|]\) is the concatenated vector of \(u\), \(v\) and their absolute difference \(|u-v|\). QR-BERT was pre-trained using crossentropy loss and Adam optimizer.

The model is then trained on a set of query-passage examples. For evidence
retrieval, the model is trained and evaluated on the FactualNLI dataset, and for answer retrieval on MSMARCO. The loss function used is as follows:

\[max_{\theta} \sum_{(q,p) \in D_{B}^{+}}{r_{\theta}(q,p) - log(\sum_{(p_i \in D_{B}}{e^{r_{\theta}(q,p_i)}})}\]
\noindent where \(D_B\) is the set of passages in a training batch \(B\), \(D^+_B\) is the set of positive query-passage pairs in \(B\) and \(r_{\theta}(q,p)\) is the inner product function between the embeddings of \(q\) and \(p\). It is trained with Adam optimizer, initial learning rate \(2 \times 10^{-5}\), batch size 256 and 10,000 warm-up steps.

QR-BERT is used to generate semantically relevant embeddings for the tokens in the queries and the passages. These tokens are used to compare the query to the passages in terms of semantics, but since the comparison between each token of the query to each token of every paragraph can become too computationally expensive, a sentence embedding is used instead of token embeddings — and so, each sentence (query or passage) embedding is calculated by averaging all the embeddings of the tokens in the sentence.

To rank each passage in terms of relevance to the query, an inner product is calculated between the query's embedding and every paragraph's embedding and the resulting scores are then used to sort the passages by most relevant. To do this they use a maximum inner product search index from the FAISS library.

In the end, the top 400 results — 200 from BM25 and 200 from QR-BERT — are re-ranked using a relevance classifier based on BERT-large. This model was fine-tuned on 20 million query-passage pairs from the MSMARCO dataset by minimizing the cross-entropy loss with Adam optimizer, initial learning rate \(2 \times 10^{-5}\) and batch size 128, on a validation set of 20,000 samples until convergence.

Each passage \(p\) is ranked according to a query \(q\)  following the function \(r\):

\[r(q,p) = softmax(W \cdot \textrm{BERT}_{\textrm{[CLS]}}([q ; p]) + b)\]

\noindent where \(W_{2 \times d}\) is a linear transformation matrix, \(d\) is the dimensionality of a BERT embedding and \(b_{2 \times 1}\) is the added bias in the transformation. The BERT embedding used is the one corresponding to the first token in the query-passage pair ([CLS]). The results are then reordered according to the new scores.

\subsection {Semantic NLP for Automated Compliance Checking}
The SNACC (Semantic Natural Language Processing - Based Automated Compliance Checking) system \cite{ZHANG2017} is able to automatically extract regulatory information from regulatory documents of the International Building Code. The regulatory information extracted is represented in First Order Logic (FOL). By representing this information using a FOL formalistm, SNACC is able to automatically check if a particular building model is consistent with regulations.

\begin{figure}
\begin{center}
\includegraphics[width=250pt]{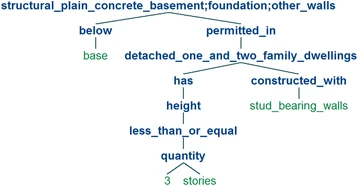}
\end{center}
\caption{Tree-based representation for the sentence: "Structural plain concrete basement, foundation or other walls below the base are permitted in detached one- and two-family dwellings three stories or less in height constructed with stud-bearing walls." \cite{ZHANG2017a}}
\label{snacc}
\end{figure}

Another interesting aspect of SNACC is that its author explored a tree-based visual representation of regulatory statements (the tree based representation was constructed from the FOL representation), and concluded that the three-based representation is easier to understand and to read by users than its FOL counterpart. Figure \ref{snacc} show an example a tree-based representation.

Although SNACC is a very advanced system in terms of automated compliance checking, its semantic analysis component is based in classical techniques of semantic information extraction, which focuses on the use of rules designed by experts. SNACC uses 146 rules for information extraction, 297 rules for semantic mappint and 9 rules for conflict resolution. The problem with this type of traditional approaches is that it is not easy to apply in a large scale to other domains, as it will usually require lots of expert knowledge in writing new rules. Nevertheless, SNACC demonstrates the potential of exploring semantic information in a legal domain.   

\subsection{Extracting Norms from Legal Text Using SRL}

Humphreys and his colleagues \cite{humphreys-etal-2020-populating} recently looked at the problem of how to automate Knowledge Extraction from legal texts, and to use that knowledge to populate legal ontologies. To address this problem, they created a system based on natural language processing and post-processing rules based on domain knowledge. Authors claim that they are the first to use a PropBank semantic role labeler to extract definitions and norms from legal text.

The system is composed by two main components. The first component is the Mate Tools semantic role labeler \cite{bjorkelund-etal-2009-multilingual}, which extracts an abstract semantic representation, as well as dependency parse trees from legal text. The second component consists in a set of rules that identify possible norms and definitions, classify their types, and map arguments in the semantic role tree to domain-specific slots in a legal ontology. The advantage of using SRL is that it enriches a sentence parse tree with relevant semantic information, which will simplify the rules that will have to be created to extract norms and definitions.

Table \ref{table:norm_extraction} shows an example of the two-staged extraction process. The modal verb 'may', identified by the semantic role labeler as the head of the sentence, triggers the activation of a rule that identifies the type of norm as a permission. The verb 'practice' is identified as the 'predicate' of the sentence, from which arguments A0 and A2 will be extracted. Another rule identifies the keywords 'to the extent' in the argument A2, which will extract a condition by including all the trigger and subsequent words until a comma, semi-colon or full stop is reached. The rightmost column shows the final norm extracted. The fields and parameters to be extracted as the result of this process were defined based on an integration of concepts and relations from the Eunomos prescriptions and Legal-URN Hohfeldian models. 

\begin{table}
\tiny
\begin{tabular}{|p{2.8cm}|p{4.0cm}|p{4.0cm}|}
\hline
\textbf{Legal text} & \textbf{Extracted SRL Roles} & \textbf{Extracted Norm} \\
\hline
A lawyer registered in a host Member State under his home-country professional title may practise as a
salaried lawyer in the employ of another lawyer, an association or firm of lawyers, or a public or private enterprise to the extent that the host Member State so permits for lawyers registered under the professional title used in that State.
&
\begin{verbatim}
<SRL>
    <PREDICATE>practice</PREDICATE>
    <A0:SBJ>
        A lawyer registered in a host 
        Member State under his 
        home-country professional title
    </A0:SBJ>
    <A2:ADV>
        practise as a salaried lawyer in
        the employ of another lawyer, an
        association or firm of lawyers,
        or a public or private enterprise,
        to the extent that the host Member
        State so permits for lawyers
        registered under the professional
        title used in that State
    </A2:ADV>
</SRL>
\end{verbatim}
&
\begin{verbatim}
<Norm>
    <NormType>Permission</NormType>
    <ActiveRole>
        A lawyer registered in a host 
        Member State under his 
        home-country professional title
    <ActiveRole>
    <Action>
        practise as a salaried lawyer in 
        the employ of another lawyer, an
        association or firm of lawyers,
        or a public or private enterprise
    </Action>
    <Condition>
        to the extent that the host Member
        State so permits for lawyers 
        registered under the professional
        title used in that State
    </Condition>
</Norm>    
\end{verbatim}
\\
\hline
\end{tabular}
\caption{Example of SRL and Norm extraction from Legal Text. Taken from \cite{humphreys-etal-2020-populating}.}
\label{table:norm_extraction}
\end{table}

To evaluate the performance of the developed system, Humphreys et al. started by determining the accuracy of the semantic role labeler on legal text, by testing it over 58 definitions and 166 norms examples. While 78.5\% of definitions had all the arguments for all the verbs correct, only 52\% of norms had fully correct arguments. Despiste this limitation, the final system managed to obtain relatively good performance when applied to the legislative text it was designed to cover, reaching a f-1 score of 76.33 in detecting Obligation type norms and 85.15 in power type norms. Additional tests were made to evaluate the system in unseen legislative body. Final results show that while some of the most important elements - Norm Type, and Active Role - are extracted with a good accuracy the system has problems with detecting passive roles and with other elements of a norm such as conditions.

%\section{Analysis}
%
%Legal Expert System vs Legal Information Retrieval vs Legal Cognitive Assistant
%
%Proactive Legal Information Retrieval and filtering accessing legal texts in a way that will simplify the process and make the retrieved data more relevant and timely.
%

\section{Conclusions}

We presented in this paper short analysis of the history and current state of the research field of AI applied to the legal domain, its promise and its challenges. We also found a couple of available products in use. We then reviewed recent trends in Deep Learning applied to NLP, Information Retrieval and Semantic Representation, which can be used to build better systems to help citizens access and understand the law.

These techniques, some so recent that its papers were published just a couple of weeks before the writing of these conclusions, can be used to help bridge the triple Natural Language Barrier facing non-professional users of a legal Information search system we described above. Of particular interest is the recent focus on systems that are able to explore the use of automatically extracted syntactic and semantic information for tasks that require natural language understanding. Semantic information - even if only provided at a shallow level - provides a deeper understanding of the text in a legal corpus, allowing us to go beyond word matching techniques. Another relevant trend is the use of weakly-supervised techniques -- with low annotation cost and easily scalable to large domains -- to automatically generate datasets of questions and answers and use them to train models in downstream tasks such as information retrieval or question answering.  

By building upon some of these techniques, we believe it will be possible to build a Proactive Legal Information Retrieval and filtering system that will permit accessing legal texts in a way that simplifies the process and makes the retrieved data more relevant and timely.

\section{Acknowledgments}
This article is a result of the project Descodificar a Legislação, supported by Competitiveness and Internationalization Operational Programme (COMPETE 2020), under the PORTUGAL 2020 Partnership Agreement, through the European Social Fund (ESF).

\bibliographystyle{splncs04.bst}
\bibliography{refs.bib}

\end{document}